\documentclass[preprint,12pt]{elsarticle}

\usepackage{amssymb}
\usepackage{amsmath}
\usepackage{fancyhdr}
\usepackage{enumitem}
\usepackage{subfigure}
\usepackage{subcaption}

\journal{Computers and Mathematics with Applications}

\begin{document}

\begin{frontmatter}



\title{ASPINN: An asymptotic strategy for solving singularly perturbed differential equations}


\author[a]{Sen Wang}
\author[a]{Peizhi Zhao}
\author[a]{Tao Song\corref{*}}

\affiliation[a]{organization={College of Computer Science and Technology},
            addressline={China University of Petroleum (East China)}, 
            city={Qingdao},
            postcode={266580}, 
            country={China}}

\cortext[*]{Corresponding author}
\ead{tsong@upc.edu.cn}

\begin{abstract}
Solving Singularly Perturbed Differential Equations (SPDEs) presents challenges due to the rapid change of their solutions at the boundary layer. In this manuscript, We propose Asymptotic Physics-Informed Neural Networks (ASPINN), a generalization of Physics-Informed Neural Networks (PINN) and General-Kindred Physics-Informed Neural Networks (GKPINN) approaches. This is a decomposition method based on the idea of asymptotic analysis. Compared to PINN, the ASPINN method has a strong fitting ability for solving SPDEs due to the placement of exponential layers at the boundary layer. Unlike GKPINN, ASPINN lessens the number of fully connected layers, thereby reducing the training cost more effectively. Moreover, ASPINN theoretically approximates the solution at the boundary layer more accurately, which accuracy is also improved compared to GKPINN. We demonstrate the effect of ASPINN by solving diverse classes of SPDEs, which clearly shows that the ASPINN method is promising in boundary layer problems. Furthermore, we introduce Chebyshev Kolmogorov-Arnold Networks (Chebyshev-KAN) instead of MLP, achieving better performance in various experiments.
\end{abstract}








\end{frontmatter}



\section{Introduction}
\label{sec1}

Singularly perturbed differential equations (SPDEs) are essential mathematical models for a wide range of physical phenomena, particularly in the study of fluid mechanics\citep{1,2}. SPDEs differ from traditional differential equations in that the former has a small positive parameter $\varepsilon$ before the highest derivative. This $\varepsilon$ is usually referred to as the perturbation parameter. As $\varepsilon$ approaches zero, these equations yield solutions that change rapidly at boundary layers. Traditionally numerical methods such as finite element method (FEM) and finite difference method (FDM) are commonly used to solve partial differential equations (PDEs). However, traditional numerical methods that use a uniform grid may fail to accurately capture regions with rapid changes in the solution, and produce large errors at the peak or boundary layer of the solution. Consequently, addressing the challenges posed by SPDE presents a significant endeavor.

With the emergence and development of deep learning, people are interested in using artificial neural networks (ANN) to solve partial differential equations\citep{3}. Operator computing methods such as FNO\citep{4} and DeepONet\citep{5} have also received a lot of attention in recent years due to their ability to learn operators. Meanwhile, PINN has become a universal deep-learning method for solving partial differential equations in the field of physics-informed machine learning. PINN takes advantage of the neural network to approximate the solution of the desired PDE and incorporates the residual term into the loss function\citep{6,7}, training and optimizing the network by minimizing the residual of the PDE. In recent years, PINN has addressed the optimization problem of neural networks through domain decomposition\citep{8,9,10,11}, model reparameterization\citep{12,13}, feature mapping\citep{14,15,16,17,18}, sequential learning\citep{19,20}, and adaptive activation functions\citep{21}. On the other hand, advanced algorithms such as loss reweighting schemes achieve a similar goal, adjusting the weight of the loss during training\citep{22,23,24,25,26,27,28}. Employing these methodologies, PINN has demonstrated a robust capacity for predictive potential. However, PINN still encounters challenges when addressing SPDEs. When $\varepsilon$ is small enough, PINN can hardly adapt to the sharp change of the solution of the equation at the boundary layer. 

This paper introduces a novel approach called ASPINN to solve SPDEs using asymptotic analysis. In common with GKPINN\citep{29}, the ASPINN approach integrates the prior knowledge from asymptotic analysis into the neural network and theoretically decomposes SPDEs into smooth, layered components, enabling separate implicit learning. Compared to GKPINN, ASPINN further simplifies the network architecture. On the basis of reducing the training cost, the prediction accuracy is improved. Our contributions are threefold:

\begin{enumerate}[label=\alph*)]
\item We propose a novel framework, ASPINN, that seamlessly amalgamates the principles of asymptotic analysis with the paradigm of PINN. This approach aims to effectively fit the steep gradient at the boundary layer when solving the SPDEs.
\item Our approach uses Chebyshev-KAN as an alternative to MLP. This purpose is to improve the function approximation performance of the ASPINN method. 
\item Our experiments prove that ASPINN is superior to GKPINN in both training cost and accuracy, providing compelling evidence for the efficacy of the proposed method. Meanwhile, Chebyshev-KAN shows better performance than MLP in the approximation function.
\end{enumerate}

The article is organized as follows. 
Section \ref{sec2} presents an overview of the PINN framework, residual-based attention and Chebyshev-KAN. In Section \ref{sec3}, we introduce singularly perturbed differential equations and discuss the problems encountered by PINN in solving SPDEs. Section \ref{sec4} proposes ASPINN and does theoretical analysis in different environments. In Section \ref{sec5}, we provide numerical examples to demonstrate the effectiveness of the proposed approach for SPDEs and compare Chebyshev-KAN with MLP in experiments. Finally, we present the conclusions of the article in Section \ref{sec6}.

\section{Related work}
\label{sec2}

In this section, we introduce the PINN structure, residual-based attention \citep{30}, and Chebyshev-KAN\citep{31,32} used in this work.  
\subsection{Physics-Informed Neural Networks}
\label{subsec2.1}

We provide a brief overview of Physics-Informed Neural Networks (PINN) in the context of solving PDEs and express a singularly perturbed differential equation as an example:
\begin{equation}
\label{eq1} 
u_t - \varepsilon u_{xx} + u_x - 3u = 0, \, x,t\in (0,1)
\end{equation}

The initial and boundary conditions are as follows:
\begin{equation} 
\label{eq2} 
u(x,0) = cos(\pi x)
\end{equation}
\begin{equation} 
\label{eq3} 
u(0,t) = 1,u(1,t) = 0
\end{equation}

The PINN solves this problem by modeling the solution with a deep neural network, denoted by $u_{\theta}(x,t)$. Where $\theta$ represents all adjustable parameter weights $W$ and biases $b$ in the neural network. PDE residuals can be defined as:
\begin{equation} 
\label{eq4} 
R_\theta(x,t)=\frac{\partial u_\theta}{\partial t}(x_r,t_r)+N\left[u_\theta\right](x_r,t_r)
\end{equation}

PINN takes into account physical laws during the training process and integrates physical equations into the loss function. Thus the differential equation is explicitly encoded by minimizing the loss function as follows:
\begin{equation} 
\label{eq5} 
L(\theta)=L_{ic}(\theta)+L_{bc}(\theta)+L_r(\theta)
\end{equation}

Where $L_{ic}$,$L_{bc}$, and $L_{r}$ represent losses related to initial conditions, boundary conditions, and PDE residuals, respectively. These terms are given by:
\begin{equation} 
\label{eq6} 
L_{ic}(\theta)=\frac{w_{ic}}{N_{ic}}\sum_{i=1}^{N_{ic}}\left|u_\theta(x_{ic}^i,0)-cos(\pi x_{ic}^i)\right|^2
\end{equation}
\begin{equation} 
\label{eq7} 
L_{bc}(\theta)=\frac{w_{bc}}{N_{bc}}\sum_{i=1}^{N_{bc}}\left(\left|u_\theta(0,t_{bc}^i)-1\right|^2+\left|u_\theta(1,t_{bc}^i)\right|^2\right)
\end{equation}
\begin{equation} 
\label{eq8} 
L_r(\theta)=\frac{w_{r}}{N_r}\sum_{i=1}^{N_r}\left|R_\theta(x_r^i,t_r^i)\right|^2
\end{equation}

Where $w_{ic}$, $w_{bc}$, and $w_{r}$ are brief weights for different terms in the loss function. $N_{ic}$, $N_{bc}$, and $N_{r}$ represent the number of initial training data, boundary training data, and internal collection points, respectively. Here $(x_{ic}^i, 0)$ denotes the initial condition point used as input at $t=0$, and $(0, t_{bc}^i)$ represents the boundary condition point when $x$ is located on both sides of the boundary. Additionally, $(x_r^i, t_r^i)$ signifies the collocation point passed to the residual $R_\theta(x,t)$. 

The training of neural network weights is carried out through a variant of gradient descent (GD)\citep{33}. For each iteration, the network weight $\theta$ is updated as follows:
\begin{equation} 
\label{eq9} 
\theta^{k+1}=\theta^{k} - \eta \cdot \bigtriangledown_{\theta}L
\end{equation}

Where $\eta$ is the learning rate, and $\bigtriangledown_{\theta}L$ is the gradient of the loss function with regard to $\theta$. 

\subsection{Residual-Based Attention}
\label{subsec2.2}

One of the inherent challenges of training neural networks is that the point-by-point error of the residuals can be overlooked when calculating the whole or the mean of the residuals. This limitation often results in an incomplete capture of spatial or temporal features. One way to solve this problem is to choose a set of weighted multipliers, either global or local. The purpose of the global multiplier is to adjust the various terms in the loss function, while the local multiplier aims to balance the effects of particular collocation points. Techniques such as casual training\citep{22}, residual-based attention (RBA) weights\citep{30}, and self-adaptive weights\citep{24} have shown remarkable performances in physics-informed neural networks. RBA weights are based on the exponential weighted moving average of the residuals. This can effectively eliminate undesirable local minima or saddle points while capturing the spatial and temporal characteristics of specific problems. The update rule for the proposed residual-based multipliers for any training point $i$ on iteration $k$ is given by:
\begin{equation} 
\label{eq10} 
\alpha_i^{k+1}\gets\left(1-\eta^\ast\right)\alpha_i^k + \eta^\ast \frac{\left| e \right|}{\left|\left| e \right|\right|_\infty}, \, i\in \{0,1,...,N\}
\end{equation}

where $N$ is the number of training points, $e_i$ is the residual of the respective loss term for point $i$ and $\eta^\ast$ is a learning rate. This method can effectively limit RBA between zero to one.
\subsection{Kolmogorov-Arnold Networks}
\label{subsec2.3}

The Kolmogorov-Arnold networks (KANs)\citep{34} are a new type of neural network inspired by the Kolmogorov-Arnold Theorem(also known as the Superposition Theorem)\citep{35}. It states that any continuous multivariate function $f(x)=f(x_1,x_2,...)$ over a bounded domain can be expressed as a combination of a finite number of continuous univariate functions and a set of linear operations. Motivated by this theorem, the proposed function $f(x)$ is given by:
\begin{equation} 
\label{eq11} 
f(x) = \sum\limits_{q=0}^{2d_{in}} g_q (\sum\limits_{p=1}^{d_{in}} \psi_{p,q} (x_p))
\end{equation}

Where $f:[0,1]^{d_{in}} \to R^{d_{out}}$, $g_q$ and $\psi_{p,q}$ are continuous univariate functions, $x = (x_1,x_2,...,x_{d_{in}})$. 
 
\subsubsection{Chebyshev Polynomials}
\label{subsec2.3.1}

Chebyshev polynomials\citep{36} are orthogonal polynomials defined on the interval $[-1,1]$. They are very popular in approximation theory\citep{37} and higher-order numerical methods for computational fluid dynamics (CFD). They satisfy the recurrence relation:
\begin{equation} 
\label{eq12} 
T_0(x) = 1 = cos(0 \times acos(x))
\end{equation}
\begin{equation} 
\label{eq13} 
T_1(x) = x = cos(1 \times acos(x))
\end{equation}
\begin{equation} 
\label{eq14} 
T_2(x) = 2x^2 - 1 = cos(2 \times acos(x))
\end{equation}
\begin{equation} 
\label{eq15} 
T_n(x) = 2xT_{n-1}(x)-T_{n-2} = cos(n \times acos(x))
\end{equation}

Due to the orthogonality, uniform approximation, fast convergence, and recursive computation, Chebyshev polynomials become an excellent choice for function approximation tasks. 

\subsubsection{Chebyshev Kolmogorov-Arnold Networks}
\label{subsec2.3.2}

Chebyshev Kolmogorov-Arnold Networks (Chebyshev-KAN) is a novel method to function approximation, combining the Kolmogorov-Arnold Theorem with Chebyshev polynomials. In the Chebyshev-KAN, the target function $f(x)$ is approximated as follows:
\begin{equation} 
\label{eq16} 
\tilde{f}(x) = \sum\limits_{j=1}^{d_{in}} \sum\limits_{k=0}^{n} \Theta_{j,k} T_k(\tilde{x}_j)
\end{equation}

Where $\tilde{x} = tanh(x)$ is the normalized input tensor, $T_k(\tilde{x}_j)$ is the k-th Chebyshev polynomial evaluated at $\tilde{x}_j$, $n$ is the degree of the Chebyshev polynomials, and $\Theta \in R^{d_{in} \times d_{out} \times (n+1)}$ are the learnable coefficients for the Chebyshev interpolation. This method makes use of Chebyshev's weighted sum to approximate $f(x)$.  
\section{Problem settings}
\label{sec3}

Singularly perturbed differential equations have a small positive parameter $\varepsilon$ before the highest derivative term. As $\varepsilon$ approaches zero, the solution can exhibit significant changes within certain regions, and its derivative potentially becomes unbounded in these regions. These certain regions are known as boundary layers or thin regions, depending on their relative positions. Here is a basic one-dimensional convection-diffusion equation as an illustration:
\begin{equation} 
\label{eq17} 
\begin{cases}
-\varepsilon u_{xx} + u_x + (1+\varepsilon)u = 0, \, x\in (0,1)\\ 
u(0)=1+e^{-\frac{1+\varepsilon}{\varepsilon}},u(1)=1+e^{-1} \\
\end{cases}
\end{equation}

Where $\varepsilon$ is a minuscule positive parameter. Owing to influences of $\varepsilon$, the solution of this equation exhibits singular behavior at $x=1$, as illustrated in Figure 1.
\begin{figure}[htbp]
\centering
\includegraphics[width=1.0\columnwidth]{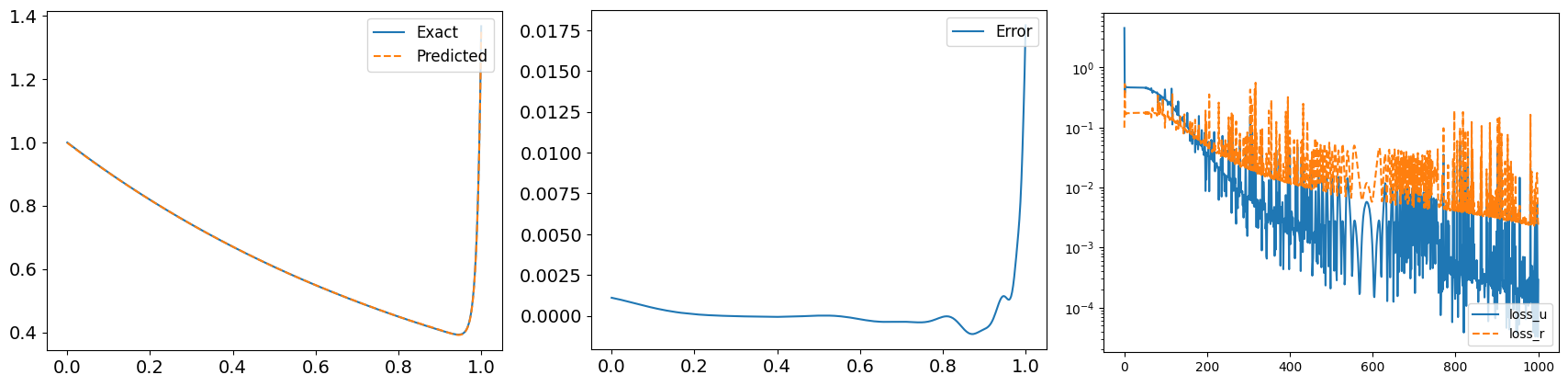} 
\caption{This is an example in Eq.(17) when $\varepsilon =0.01$. (left): Ground truth and predictions for PINN; (middle): Point-wise error for PINN; (right): Loss of PINN}
\label{fig1}
\end{figure}

Where we set $\varepsilon =0.01$ in Eq.(17) and the collocation point is 1000 points obtained by Latin hypercube sampling. This is the result of training $1.0e^5$ iterations with PINN. As you can see from the figure, PINN achieves an efficient fit.
\begin{figure}[htbp]
\centering
\includegraphics[width=1.0\columnwidth]{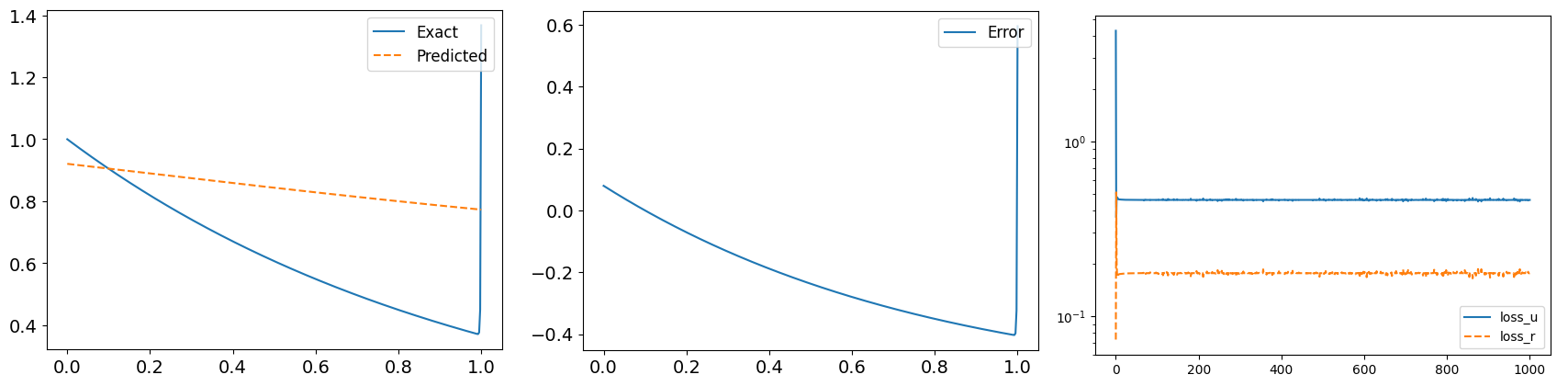} 
\caption{This is an example in Eq.(17) when $\varepsilon =0.001$. (left): Ground truth and predictions for PINN; (middle): Point-wise error for PINN; (right): Loss of PINN}
\label{fig2}
\end{figure}

However, as $\varepsilon$ continues to shrink, the solution to the equation exhibits a more pronounced variation at the boundary layer. This brings great challenges to the solving of PINN. We can see from Figure 2 that PINN is completely unable to learn steep changes at the boundary layer when $\varepsilon =0.001$. This demonstrates the limitations of traditional PINN in the face of SPDEs.

For this situation, there are studies to propose effective network architectures for SPDEs. Some researchers propose the asymptotic parameter PINN (PAPINN)\citep{38}, which approximates smooth solutions by optimizing neural networks with large perturbation parameters, which is then used as the initial value of the neural network with small perturbation parameters to approximate the singular solution. But when the perturbation parameter $\varepsilon$ becomes small enough, PAPINN needs to loop more times to adapt to changes in $\varepsilon$, which poses challenges to training efficiency and even accuracy. Then, using the prior knowledge obtained from the asymptotic analysis, the General-Kindred Physics-Informed Neural Networks (GKPINN)\citep{29} is proposed. This method sets exponential-type layers to approximate large gradient solutions and can fit any small perturbation parameters, demonstrating strong generalization in the face of SPDEs. 

\section{Method}
\label{sec4}

Our network architecture is based on asymptotic expansion\citep{39} and singular perturbation theory. For various types of SPDEs, their solutions are usually divided into two components by asymptotic expansions. One section describes how the solution behaves inside the boundary or inner layer, referred to as the layer part, while the other deals with how the solution behaves outside these areas, called the smoothing part. So we call $u_{as}$ an asymptotic expansion of order $m$. Then there is a constant $C$, and when $x \in [0,1]$ and $\varepsilon$ is small enough, the formula is as follows:
\begin{equation} 
\label{eq18} 
\begin{cases}
\left| u(x)-u_{ax} \right| \leq C\varepsilon^{m+1}\\ 
u_{as} = u_m + v_m \\
\end{cases}
\end{equation}

Where $u_m$ and $v_m$ are smoothing part and layer part, respectively. Within this manuscript, We will use the idea of asymptotic analysis\citep{40} to elucidate boundary and inner layer phenomena in various types of SPDEs.
\subsection{Ordinary Differential Equations}
\label{subsec4.1}

Let's start with a simple convection-diffusion equation:
\begin{equation} 
\label{eq19} 
\begin{cases}
\varepsilon u^{''} + b(x)u^{'} + c(x)u = f(x), \, x \in (0,1)   \\ 
u(0) = 0, \, u(1) = 1 \\
\end{cases}
\end{equation}

In this case, the position of the boundary layer is affected by the value of b(x):
\begin{equation} 
\label{eq20} 
\begin{cases}
b(x)>0 \to x=1\\ 
b(x)<0 \to x=0 \\
\end{cases}
\end{equation}

When $b(x)>0$, with the help of the matched asymptotic expansion, solution $u(x)$ has the following m-order asymptotic expansion:
\begin{equation} 
\label{eq21} 
u_{as}(x) = \sum\limits_{\alpha=1}^{m} \varepsilon^{\alpha} u_{\alpha}(x) + \sum\limits_{\beta=1}^{m} \varepsilon^{\beta} v_{\beta}(\frac{1-x}{\varepsilon})
\end{equation}

This method treats $u_0$ as a reduced solution and $v_0(\frac{1-x}{\varepsilon}) = (u(1) - u_0(1))e^{-b(1)\frac{1-x}{\varepsilon}}$, resulting in the following asymptotic expansion:
\begin{equation} 
\label{eq22} 
\left| u(x)-(u_0 + v_0) \right| \leq C\varepsilon, \, m=0
\end{equation}

Like $u_m$ and $v_m$, $v_0$ acts as the layer part and realizes the correction in the boundary layer region through the exponential layer, while $u_0$ represents the smooth part of the asymptotic expansion, capturing the smooth behavior of the region except for the boundary layer in Eq.(22). Therefore, if $b \ne 0$ when $x$ is in the domain, we have the following conclusion($x=a$ is the position of the boundary layer):
\begin{equation} 
\label{eq23} 
u_{as} = u_0(x) + (u(a) - u_0(a))e^{-b(a)\frac{a-x}{\varepsilon}}
\end{equation}

\subsection{Partial Differential Equations}
\label{subsec4.2}

First consider elliptic partial differential equations in the space domain, which are often used to describe steady-state problems. We define the mathematical set $\Omega = (0,1 )^2 $ and the standard format is shown as below:
\begin{equation} 
\label{eq24} 
\begin{cases}
-\varepsilon \Delta u + b(x,y)\nabla u + c(x,y)u = f(x,y), \, in \, \Omega \\ 
u(x,y)=0, \, on \, \partial \Omega \\
\end{cases}
\end{equation}

In this equation, $b_1$ and $b_2$ in $b = b(x,y) = (b_1,b_2)$ together affect the position of the boundary layer. We only consider the case where $b_1=0$ or $b_2=0$ in this manuscript. Thus, under the assumption of $b_1>0$, the asymptotic expansion of $u$ is formulated as:
\begin{equation} 
\label{eq25} 
u_{as}(x,y) = u_0(x,y) + (u(1,y) - u_0(1,y))e^{-b(1,y)\frac{1-x}{\varepsilon}}
\end{equation}

For parabolic partial differential equations in the space-time domain $Q = \left(0, 1\right) \times (0, T]$, we have the following equation:
\begin{equation} 
\label{eq26} 
\begin{cases}
u_t -\varepsilon u_{xx} + b(x,t)u_x + c(x,t)u = f(x,t),  (x,t)\in Q \\ 
u(x,0) = g(x) \\
u(0,t) = q_0(t),u(1,t) = q_1(t)
\end{cases}
\end{equation}

For fixed $t>0$, the treatment of the boundary layer is very similar to the ordinary differential equation in Eq.(19). In cases when $b(x,t)>0$, the solution $u(x,t)$ usually shows a boundary layer at $x=1$ and can be incrementally decomposed into:
\begin{equation} 
\label{eq27} 
u(x,t) = u_0(x,t) + (u(1,t) - u_0(1,t))e^{-b(1,t)\frac{1-x}{\varepsilon}}
\end{equation}

\subsection{Network Structure}
\label{subsec4.3}

Above, we introduced diverse classes of SPDEs in ordinary differential equations and partial differential equations. The solutions of equations and their asymptotic expansions can be divided into two parts: the smooth part and the layer part. The layer part shows exponential characteristics and can be represented by the exponential function. GKPINN is a novel architecture based on prior knowledge of layer location, the model can be succinctly represented by the equation:
\begin{equation} 
\label{eq28} 
\mathit{GKPINN} = u_0 + \sum\limits_{i=1}^{N} u_i \ast exp(-\alpha_i))
\end{equation}

Where $u_0$ represents the reduced solution, fitted by a shallow neural network. $N$ indicates the number of boundary layers, and $exp(-\alpha_i)$ refers to the exponential layers handled by different boundary layers, which vary according to their location and type. $u_i$ represents the neural network combined with the exponential layer $exp(-\alpha_i)$.

Based on GKPINN, we further explore the prior knowledge in SPDEs. $u_i$ can be represented by $u_0$ according to the asymptotic expansion of the solution $u$. This insight prompts the construction of the ASPINN illustrated in Figure 3. The method not only reduces the complexity of the model but also makes the fitting of the predicted solutions more accurate. Therefore, our network structure is given by:
\begin{equation} 
\label{eq29} 
\mathit{ASPINN} = u_0 + \sum\limits_{i=1}^{N} (u(a_i) - u_0(a_i)) \ast exp(-b(a_i)\frac{a_i-x}{\varepsilon})
\end{equation}

\begin{figure*}[t]
\centering
\includegraphics[width=0.7\textwidth]{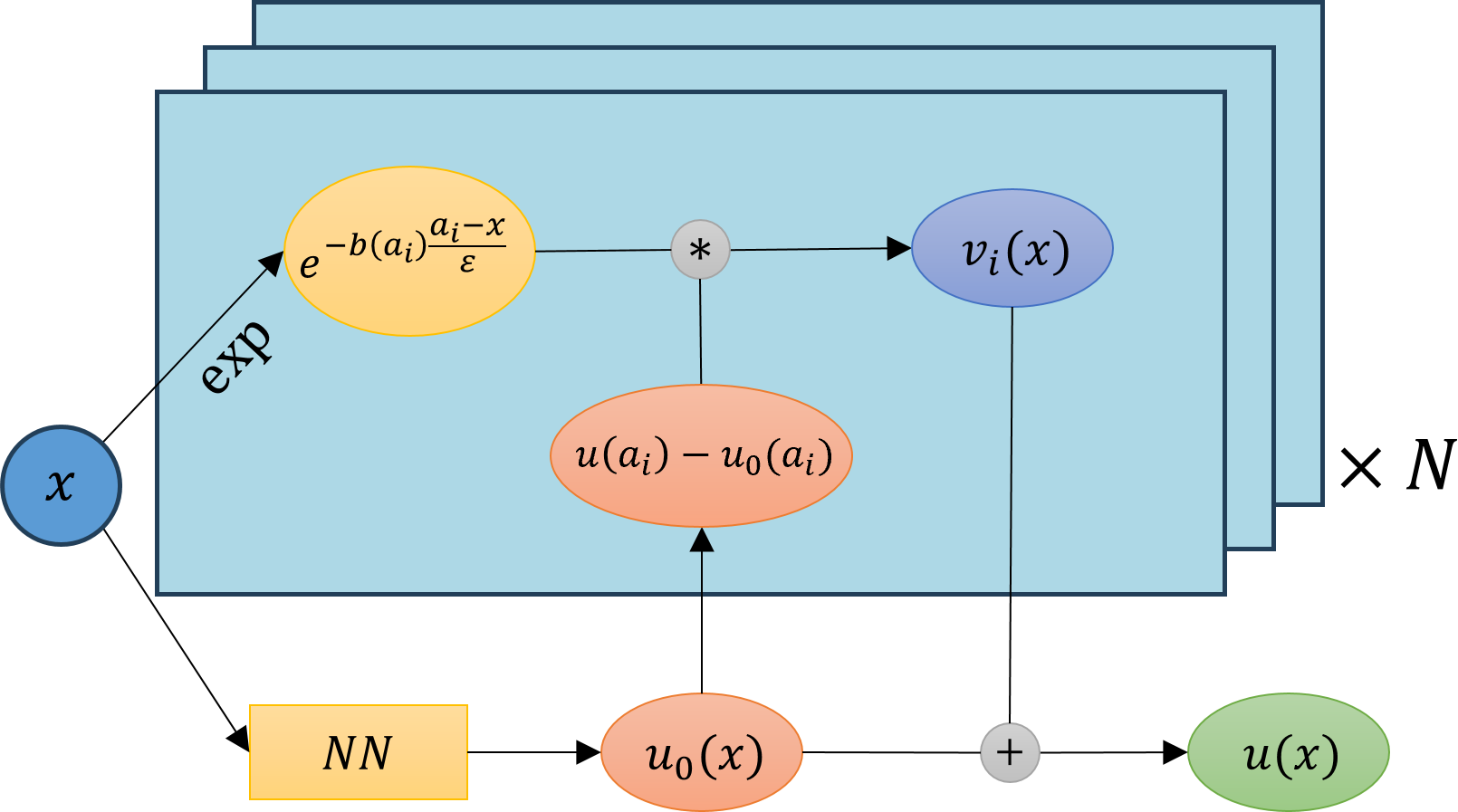} 
\caption{The architecture of ASPINN. $x$ and $u(x)$ represent the input of the model and solution of the problem, respectively. $NN$ represents the shallow neural network, $a_i$ denotes the position of the boundary layer, $b(x)$ represents the coefficient of $u_x$, and $exp$ signifies the exponential operation.}
\label{fig3}
\end{figure*}
\section{Numerical experiments}
\label{sec5}

This section applies ASPINN to various kinds of SPDEs, including both ordinary and partial differential equations. In the following examples, We compare experimental results with those obtained using PINN and GKPINN.

Thereafter, we provide a detailed description of the parameter settings used in experiments. The selected perturbation parameter $\varepsilon$ is $1 \times 10^{-3}$. MLP has two hidden layers with 100 neurons, Chebyshev-KAN has one hidden layer with 8 neurons and the degree of the Chebyshev polynomials $n=5$. We uniformly used ADAM as the optimizer with a learning rate of $0.001$. Except for MLP, which used the Sigmoid activation function in the one-dimensional experiment, the rest of the experiments used the Tanh activation function. We set $w_{ic} = w_{bc} =  w_{r} = 1$ to ensure that the loss function is unbiased. The test set of ODE is generated by analytic solution, and the test set of PDE is obtained by using high-precision finite difference methods\citep{2,41}. We initialize the RBA weights to $1$ and update them with a learning rate of $\eta^\ast=0.0001$, as described in Eq. (10).

For each experiment, $1.0e^5$ iterations were uniformly trained. We evaluate model performance in terms of total measured training time and relative $L_2$:
\begin{equation} 
\label{eq30} 
L_2 = \left|\left| \hat{u}-u \right|\right|_2 = \sqrt{\frac{\sum_{i=1}^{N_{test}}{\left| \hat{u}(x^i)-u(x^i) \right|}^2}{\sum_{i=1}^{N_{test}}{\left| \hat{u}(x^i) \right|}^2}}
\end{equation}

Where $N_{test}$ represents the number of points in the test set, and $\hat{u}$ denotes the approximate solution obtained through deep learning. All these measurements were conducted on an Nvidia GeForce RTX-3090 GPU.
\subsection{Ordinary Differential Equations}
\label{subsec5.1}

For one-dimensional equations, collect 1000 collocation points through Latin hypercube sampling. We shall examine the boundary layer on different sides and begin with the following question:
\subsubsection{Example 1}
\label{subsec5.1.1}
\begin{equation} 
\label{eq31} 
\begin{cases}
-\varepsilon u_{xx} + u_x = \varepsilon \pi^2 sin(\pi x) + \pi cos(\pi x), \, x\in (0,1) \\ 
u(0)=0, \, u(1)=1 \\
\end{cases}
\end{equation}

There exists an analytical solution to this problem:
\begin{equation} 
\label{eq32} 
u(x)=sin(\pi x) + \frac{e^{\frac{x}{\varepsilon}}-1}{e^{\frac{1}{\varepsilon}}-1}
\end{equation}

In this equation, we could find that $b(x)=1>0$, and thus this problem's solutions feature an exponential boundary layer at $x = 1$. Based on our prior knowledge, the asymptotic expansion of $u$ can be expressed as:
\begin{equation} 
\label{eq33} 
u_{as}(x) = u_0(x) + (1 - u_0(1))e^{-\frac{1-x}{\varepsilon}}
\end{equation}
\begin{table*}[htbp]
\centering
\includegraphics[width=0.5\columnwidth]{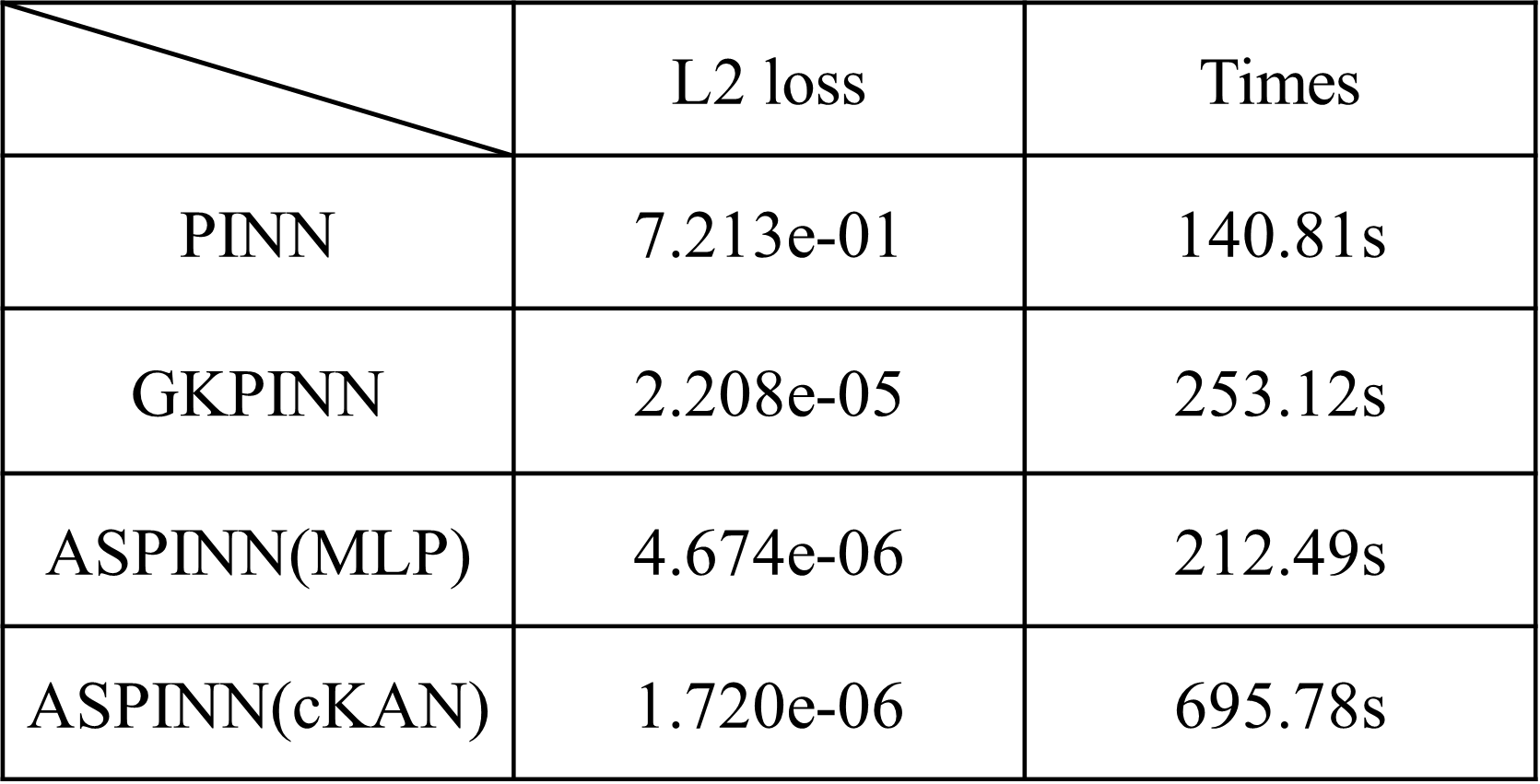} 
\caption{Relative $L_2$ and computational time comparison between different models and training strategies in Eq. (31). Time is measured on Nvidia GeForce RTX-3090 GPU}
\label{table1}
\end{table*}

We build the network architecture of GKPINN and ASPINN according to the asymptotic expansion, and use Chebyshev-KAN for this experiment. Figure 4 shows the prediction result, the numerical error, and the loss error plots for different cases. It can be seen that the solutions of the equations are effectively fitted in all three cases. We compare the three numerical errors and find that the error of ASPINN using Chebyshev-KAN is smaller than the other two. Table 1 shows that both GKPINN and ASPINN effectively approximate the solution to the problem.  Among them, ASPINN not only reduces the training cost by about $20\%$ compared to GKPINN, but also achieves a significant improvement in accuracy. Meanwhile, Chebyshev-KAN demonstrated superior performance, with a $63.2\%$ improvement compared to the MLP.

\begin{figure}[t]
    \centering
    \begin{minipage}{1.0\textwidth}
        \centering
        \includegraphics[width=1.0\textwidth]{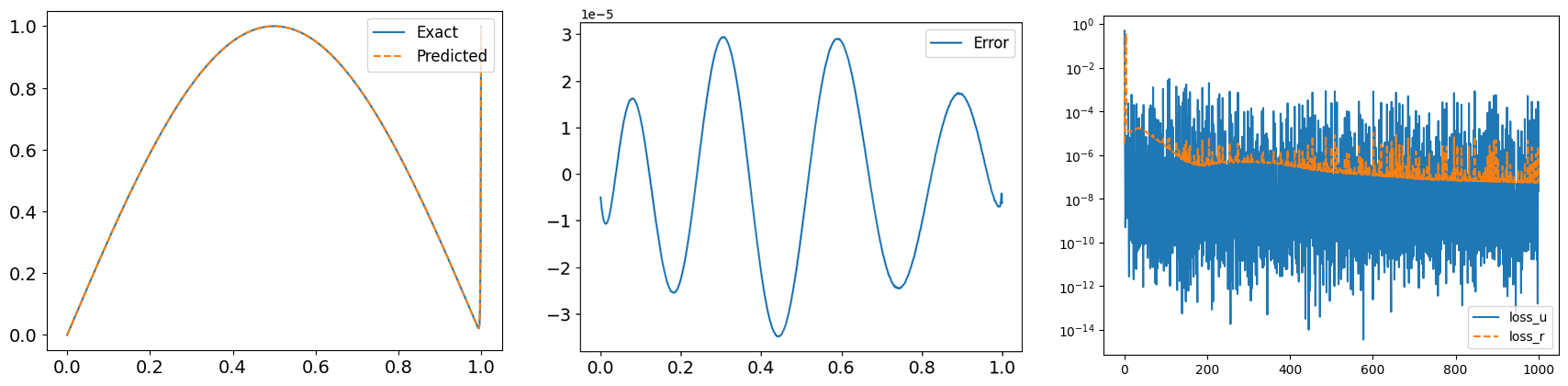}  
        \captionof{subfigure}{GKPINN(MLP)}
    \end{minipage}
    
    
    \begin{minipage}{1.0\textwidth}
        \centering
        \includegraphics[width=1.0\textwidth]{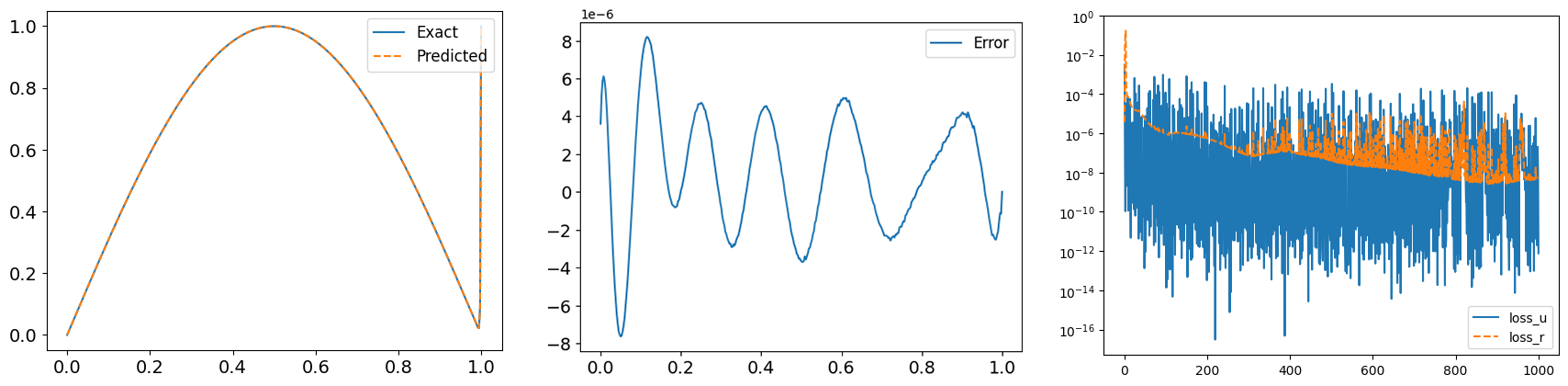} 
        \captionof{subfigure}{ASPINN(MLP)}
    \end{minipage}

    
    \begin{minipage}{1.0\textwidth}
        \centering
        \includegraphics[width=1.0\textwidth]{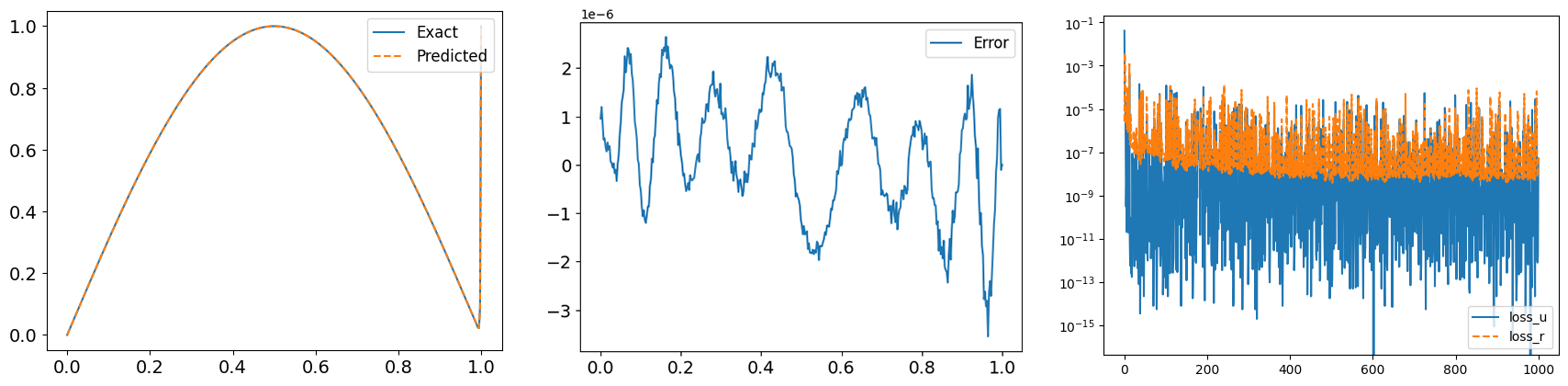}   
        \captionof{subfigure}{ASPINN(Chebyshev-KAN)}
    \end{minipage}

    \caption{Solution profile (Left), numerical error (middle), and loss error (right)}
    \label{fig4}
\end{figure}

\begin{figure}[t]
    \centering
    \begin{minipage}{1.0\textwidth}
        \centering
        \includegraphics[width=1.0\textwidth]{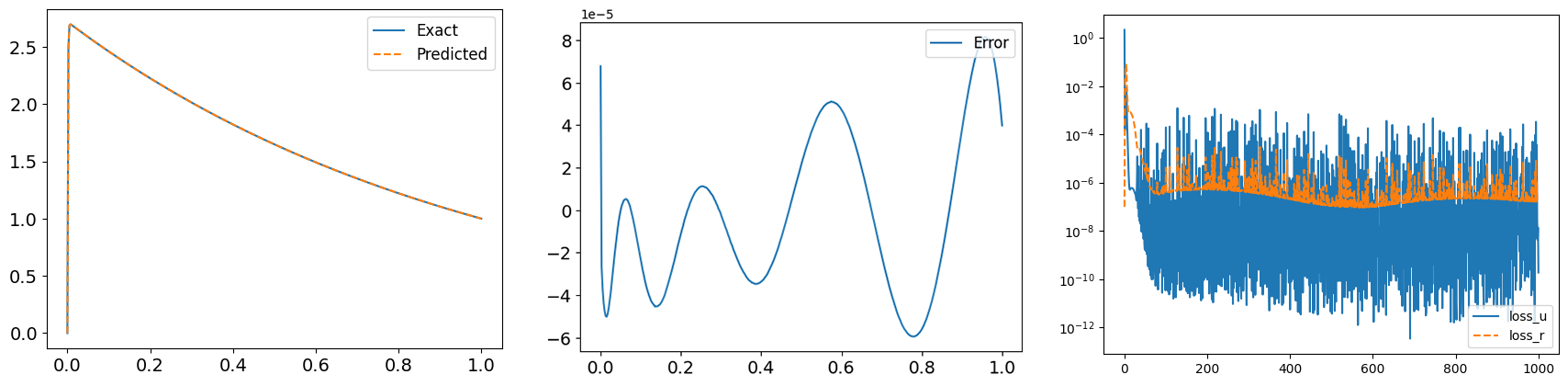}  
        \captionof{subfigure}{GKPINN(MLP)}
    \end{minipage}
    
    
    \begin{minipage}{1.0\textwidth}
        \centering
        \includegraphics[width=1.0\textwidth]{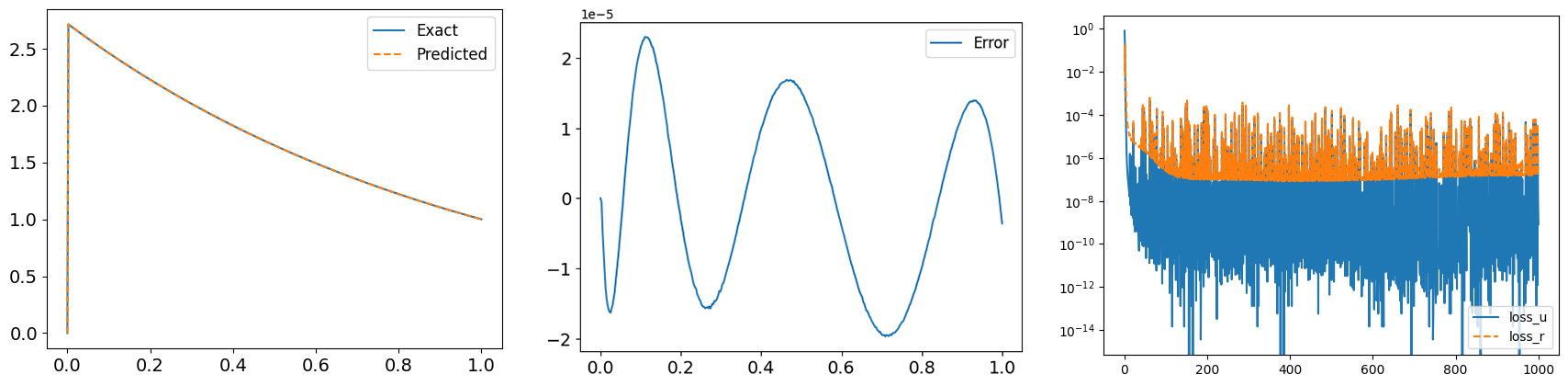} 
        \captionof{subfigure}{ASPINN(MLP)}
    \end{minipage}

    
    \begin{minipage}{1.0\textwidth}
        \centering
        \includegraphics[width=1.0\textwidth]{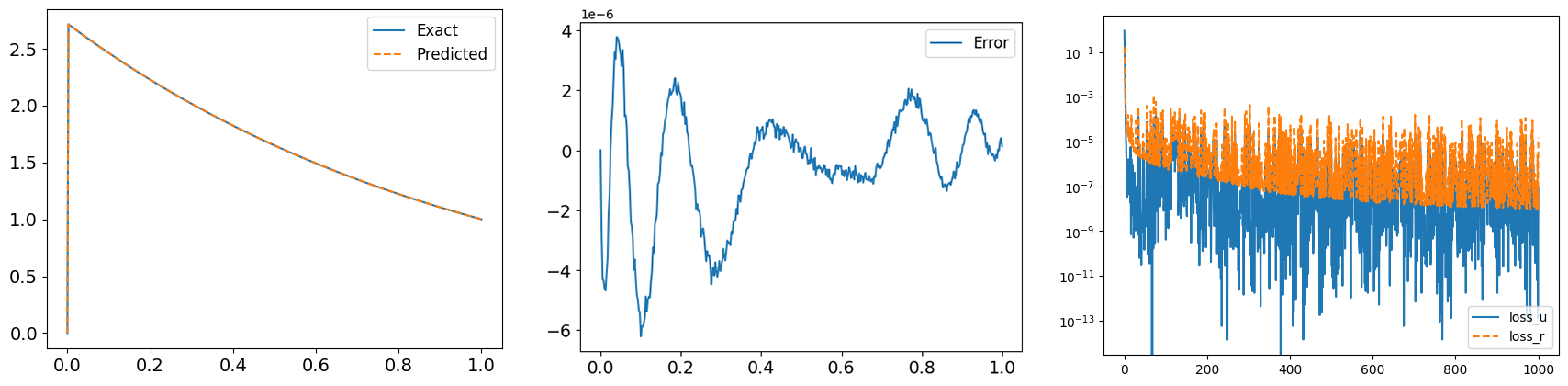}   
        \captionof{subfigure}{ASPINN(Chebyshev-KAN)}
    \end{minipage}

    \caption{Solution profile (Left), numerical error (middle), and loss error (right)}
    \label{fig5}
\end{figure}

\subsubsection{Example 2}
\label{subsec5.1.2}
Next we consider the case that boundary layer $x=0$:
\begin{equation} 
\label{eq34} 
\begin{cases}
\varepsilon u_{xx} + (1+\varepsilon)u_x + u = 0, \, x\in (0,1)\\ 
u(0)=0, \, u(1)=1 \\
\end{cases}
\end{equation}

There exists an analytical solution to this problem:
\begin{equation} 
\label{eq35} 
u(x)=\frac{e^{-x}-e^{-\frac{x}{\varepsilon}}}{e^{-x}-e^{-\frac{1}{\varepsilon}}}
\end{equation}

Based on the prior knowledge in this equation, we find that $b(x)=-1<0$, and the position of the boundary layer is the opposite of Example 1. The asymptotic expansion $u_{as}$ is given by:
\begin{equation} 
\label{eq36} 
u_{as}(x) = u_0(x) + (0 - u_0(0))e^{\frac{-x}{\varepsilon}}
\end{equation}

The results of this experiment are shown in Figure 5. We can see the good performance of ASPINN with Chebyshev-KAN from numerical error and loss error. The results for all methods are detailed in Table 2:
\begin{table*}[htbp]
\centering
\includegraphics[width=0.5\columnwidth]{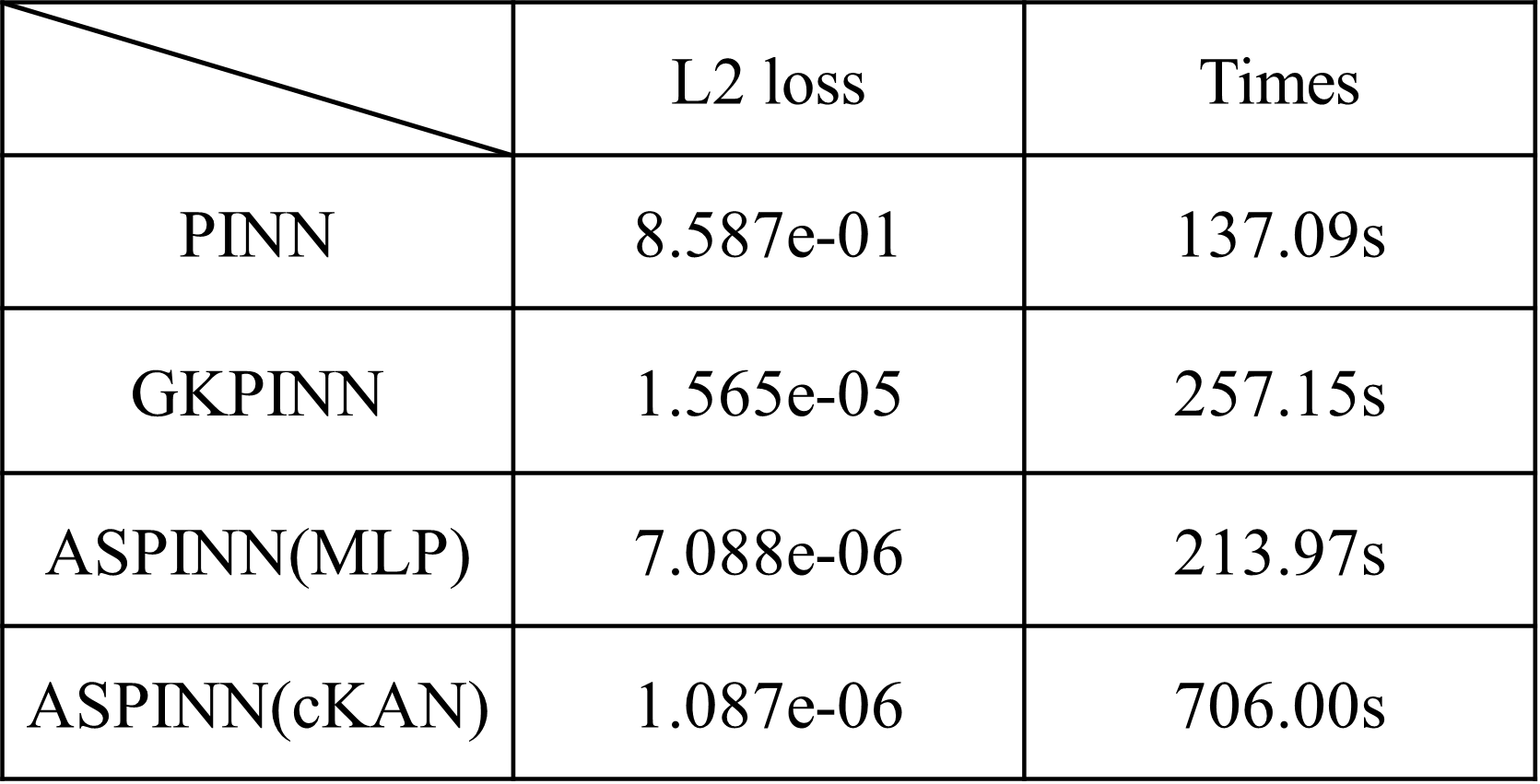} 
\caption{Relative $L_2$ and computational time comparison between different models and training strategies in Eq. (34). Time is measured on Nvidia GeForce RTX-3090 GPU}
\label{table2}
\end{table*}

According to the results of the table, we conclude that ASPINN improves both training efficiency and accuracy in ODEs, while Chebyshev-KAN greatly optimizes the accuracy compared to MLP.

\subsection{Ordinary Differential Equations}
\label{subsec5.2}

For two-dimensional and time-varying equations, we employ Latin hypercube sampling to select 10000 collocation points and randomly select 100 points each from the initial and boundary. The test set is obtained by employing high-precision finite difference methods. 
\subsubsection{Example 3}
\label{subsec5.2.1}
\begin{equation} 
\label{eq37} 
\begin{cases}
-\varepsilon (u_{xx} + u_{yy}) + u_x = 0, \, (x,y) \in (0,1)^2 \\ 
u(x,0) = u(x,1) = 0 \\
u(0,y) = sin(\pi y),u(1,y) = 2sin(\pi y) \\
\end{cases}
\end{equation}

In this equation, we note that $b = b(x,y) = (b_1,b_2) = (1,0)$, so we can infer that the boundary layer is located at $x=1$, which leads us to employ $e^{-\frac{1-x}{\varepsilon}}$ as the exponential layer. The asymptotic expansion of $u$ is as follows:
\begin{equation} 
\label{eq38} 
u_{as}(x,y) = u_0(x,y) + (2sin(\pi y) - u_0(1,y))e^{-\frac{1-x}{\varepsilon}}
\end{equation}

\begin{figure}[htbp]
    \centering
    \begin{minipage}{1.0\textwidth}
        \centering
        \includegraphics[width=1.0\textwidth]{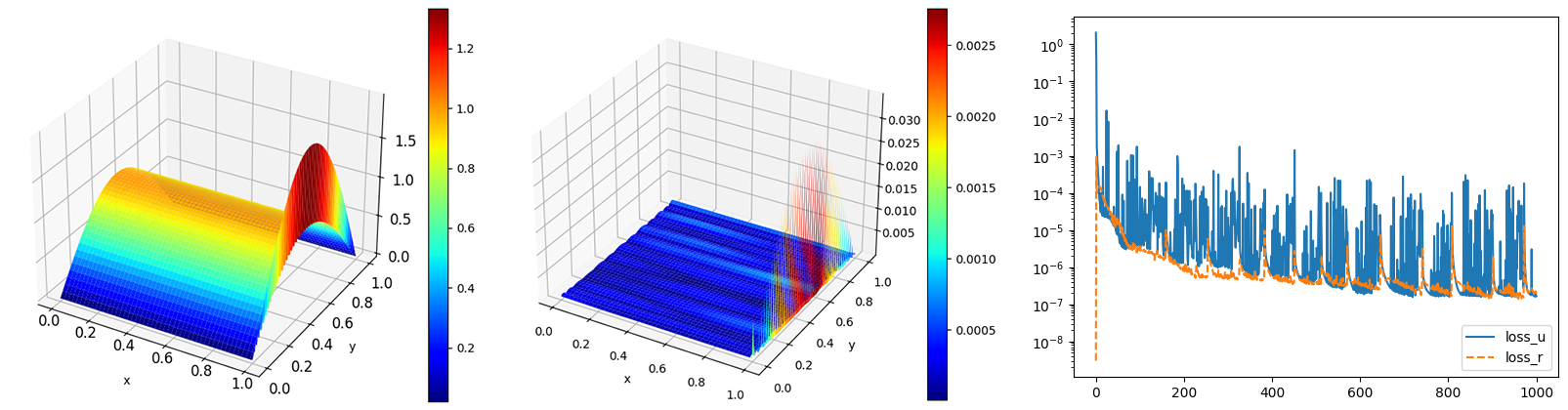}  
        \captionof{subfigure}{GKPINN(MLP)}
    \end{minipage}
    
    
    \begin{minipage}{1.0\textwidth}
        \centering
        \includegraphics[width=1.0\textwidth]{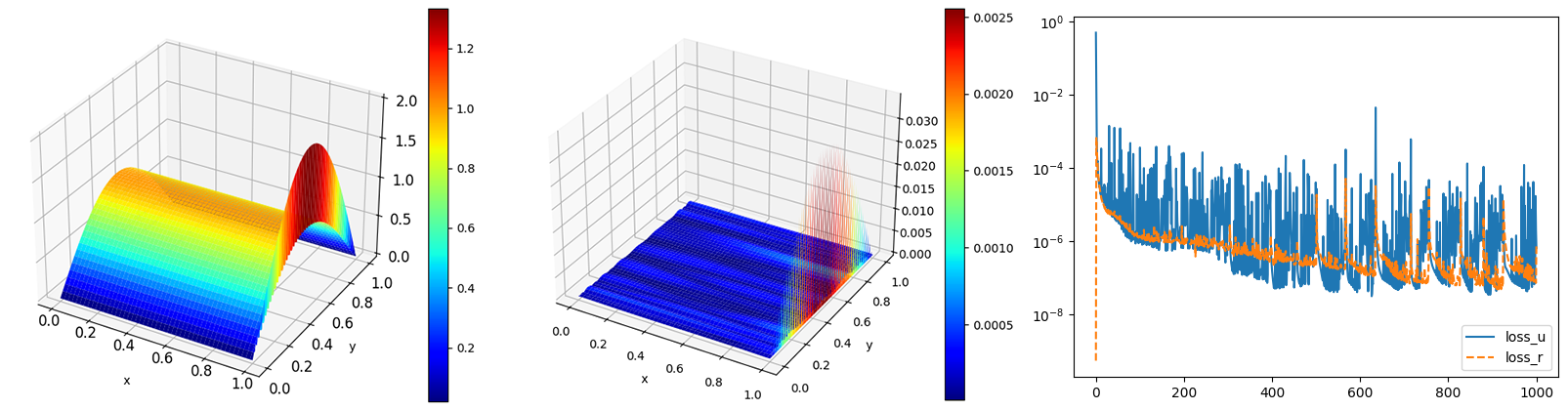} 
        \captionof{subfigure}{ASPINN(MLP)}
    \end{minipage}

    
    \begin{minipage}{1.0\textwidth}
        \centering
        \includegraphics[width=1.0\textwidth]{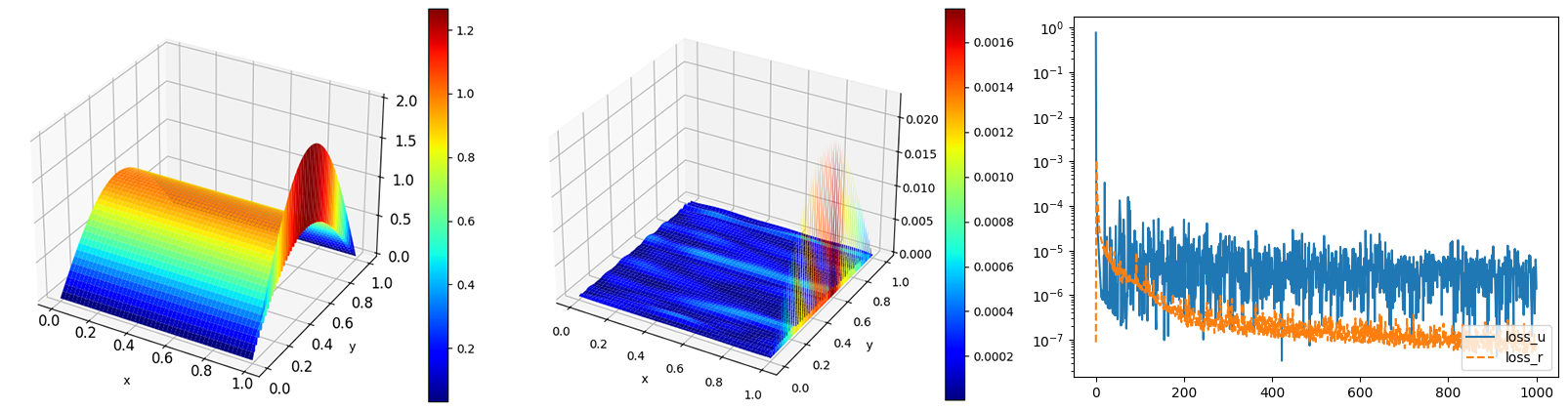}   
        \captionof{subfigure}{ASPINN(Chebyshev-KAN)}
    \end{minipage}

    \caption{Solution profile (Left), numerical error (middle), and loss error (right)}
    \label{fig6}
\end{figure}
\begin{table*}[htbp]
\centering
\includegraphics[width=0.5\columnwidth]{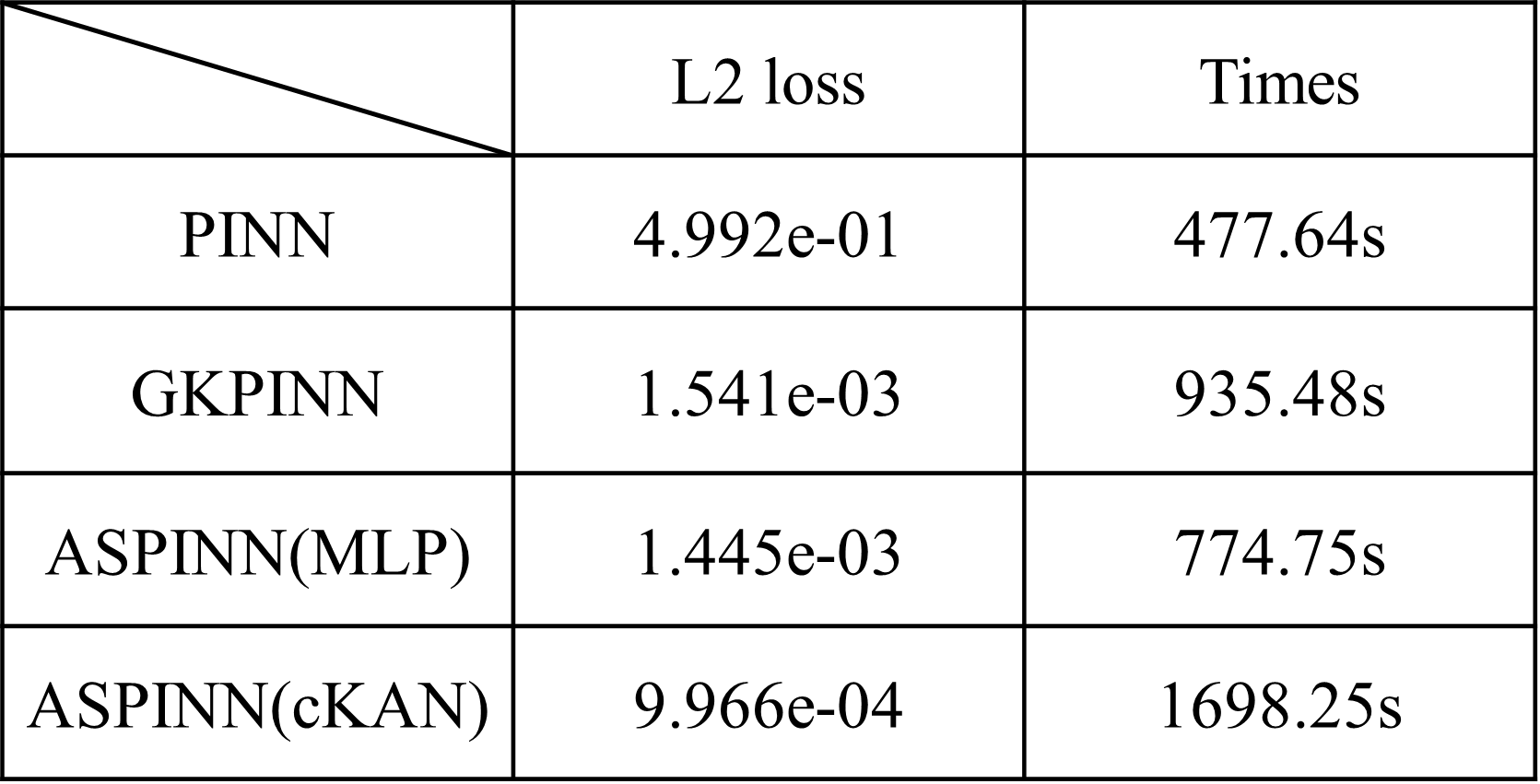} 
\caption{Relative $L_2$ and computational time comparison between different models and training strategies in Eq. (37). Time is measured on Nvidia GeForce RTX-3090 GPU}
\label{table3}
\end{table*}
As shown in Figure 6, we can see from the solution profile that the solution of the equation has a nearly vertical steep climb at $x=1$. Meanwhile, ASPINN's results are more favorable with the help of Chebyshev-KAN. More precise results are shown in Table 4:
\subsubsection{Example 4}
\label{subsec5.2.2}
\begin{equation} 
\label{eq39} 
\begin{cases}
\varepsilon (u_{xx} + u_{yy}) + u_y = 0, \, (x,y) \in (0,1)^2 \\ 
u(0,y) = u(1,y) = 0 \\
u(x,0) = 2sin(\pi x),u(x,1) = sin(\pi x) \\
\end{cases}
\end{equation}

Next we consider the case of different boundary layer locations in different dimensions. The boundary layer for this equation is present at $y=0$, while the asymptotic expansion $u_{as}$ can be concluded as:
\begin{equation} 
\label{eq40} 
u_{as}(x,y) = u_0(x,y) + (2sin(\pi x) - u_0(x,0))e^{\frac{-y}{\varepsilon}}
\end{equation}
\begin{table*}[htbp]
\centering
\includegraphics[width=0.5\columnwidth]{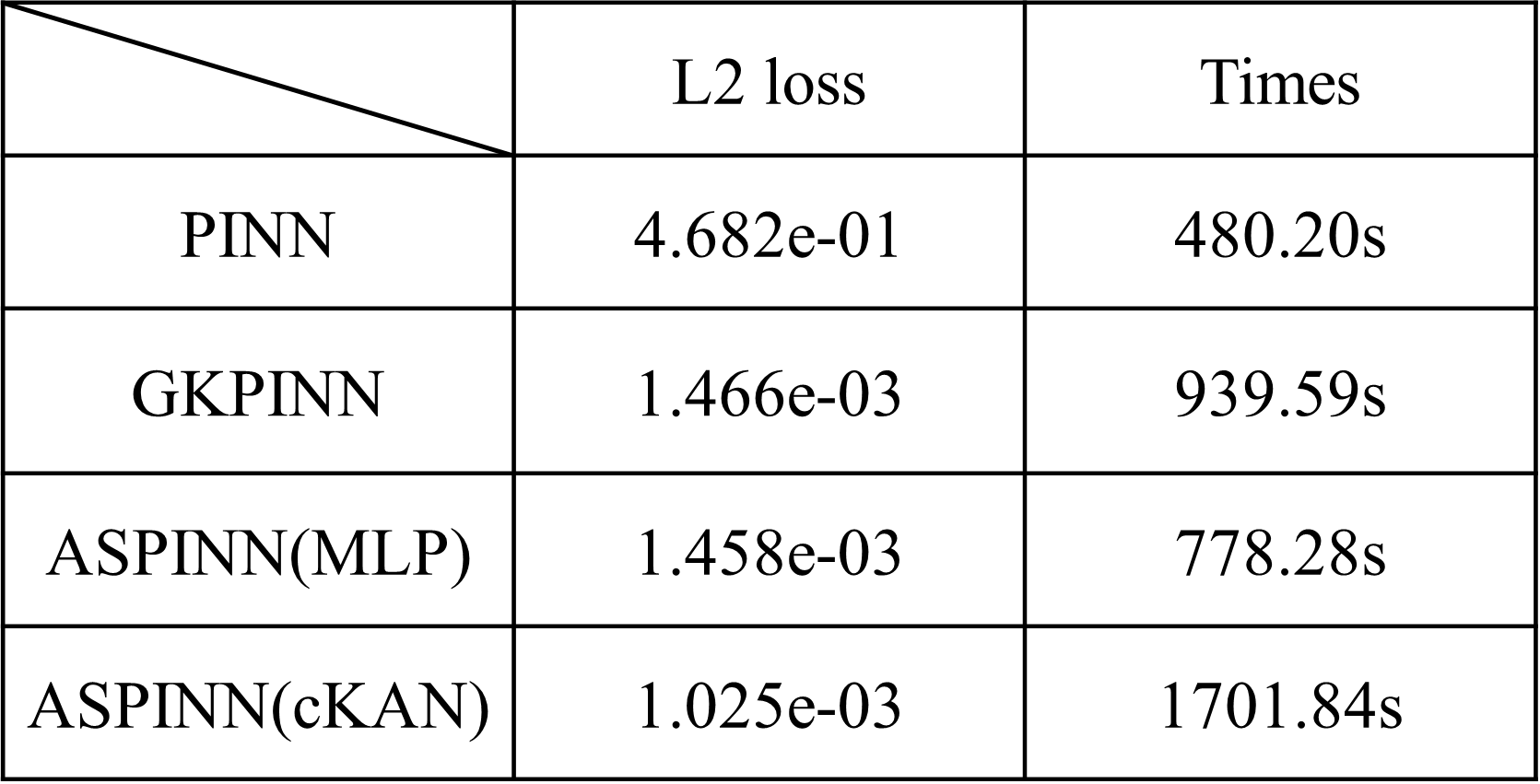} 
\caption{Relative $L_2$ and computational time comparison between different models and training strategies in Eq. (39). Time is measured on Nvidia GeForce RTX-3090 GPU}
\label{table4}
\end{table*}

The surfaces of the numerical solutions, the numerical error, and the loss error plots for the three cases are shown in Figure 7. The results for the methods are detailed in Table 4. We can come to the same conclusion as in example 3: ASPINN is significantly faster than GKPINN and the best-performing model is ASPINN (Chebyshev-KAN). 
\subsubsection{Example 5}
\label{subsec5.2.3}

Finally, we will examine the differential equations in the time domain:
\begin{equation} 
\label{eq41} 
\begin{cases}
u_t - \varepsilon u_{xx} - u_x -u = 0, \, (x,t) \in (0,1) \times (0,1] \\ 
u(x,0) = cos(2\pi x) \\
u(0,t) = 0,u(1,t) = 1 \\
\end{cases}
\end{equation}

For this kind of problem, we have $b = b(x,t) = -1 < 0$ and thus this problem's solutions feature an exponential boundary layer at $x = 0$.

\begin{figure}[htbp]
    \centering
    \begin{minipage}{1.0\textwidth}
        \centering
        \includegraphics[width=1.0\textwidth]{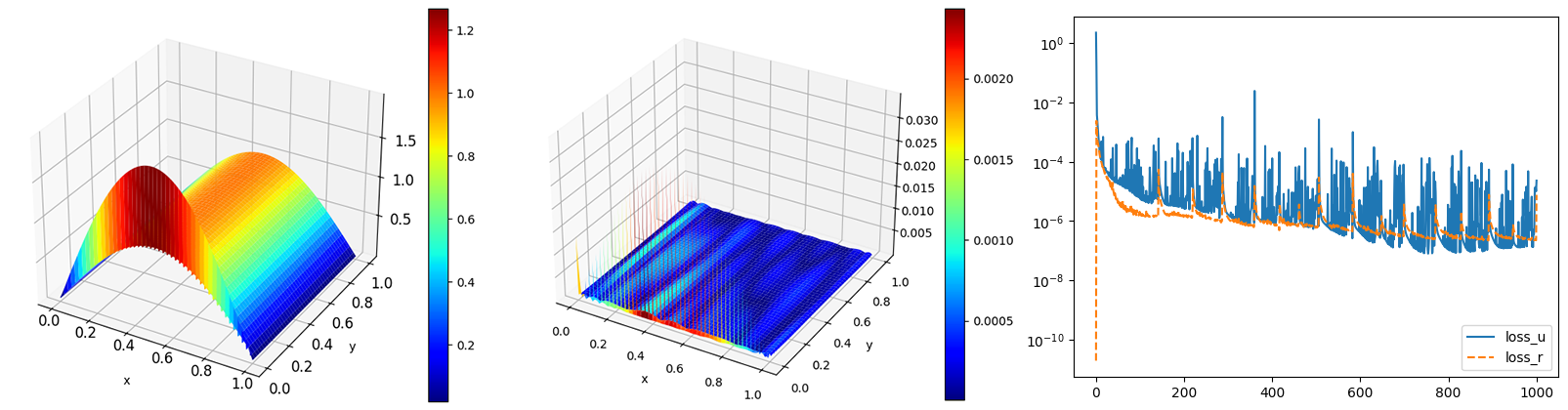}  
        \captionof{subfigure}{GKPINN(MLP)}
    \end{minipage}
    
    
    \begin{minipage}{1.0\textwidth}
        \centering
        \includegraphics[width=1.0\textwidth]{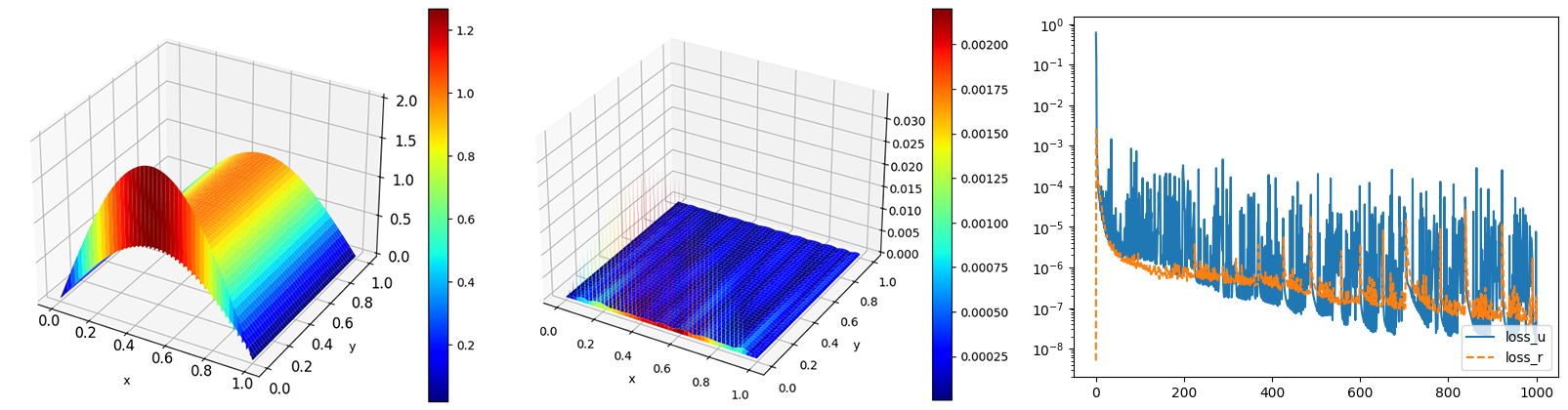} 
        \captionof{subfigure}{ASPINN(MLP)}
    \end{minipage}

    
    \begin{minipage}{1.0\textwidth}
        \centering
        \includegraphics[width=1.0\textwidth]{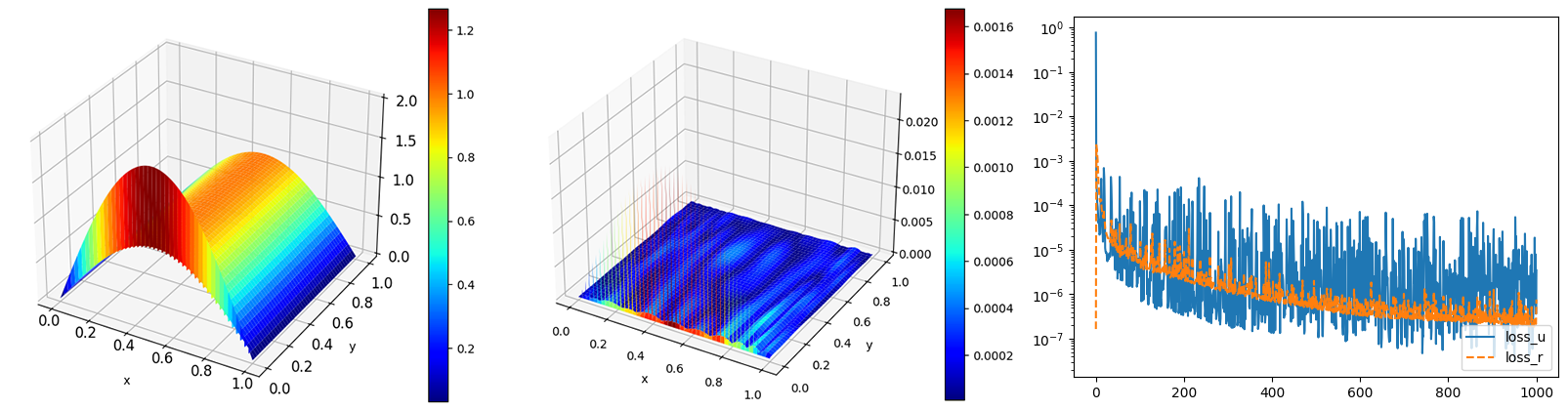}   
        \captionof{subfigure}{ASPINN(Chebyshev-KAN)}
    \end{minipage}

    \caption{Solution profile (Left), numerical error (middle), and loss error (right)}
    \label{fig7}
\end{figure}

\begin{figure}[htbp]
    \centering
    \begin{minipage}{1.0\textwidth}
        \centering
        \includegraphics[width=1.0\textwidth]{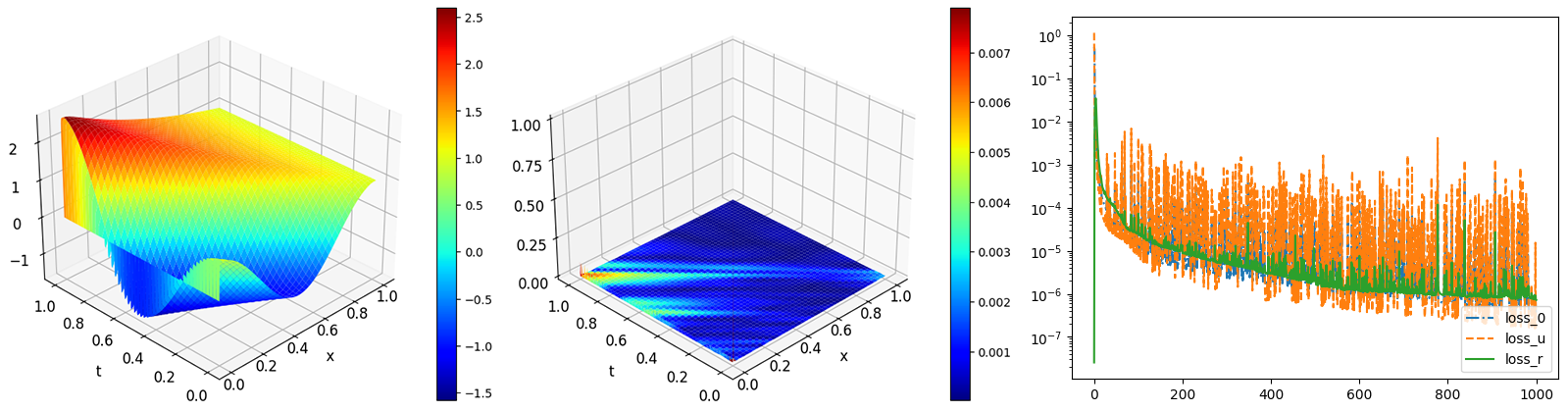}  
        \captionof{subfigure}{GKPINN(MLP)}
    \end{minipage}
    
    
    \begin{minipage}{1.0\textwidth}
        \centering
        \includegraphics[width=1.0\textwidth]{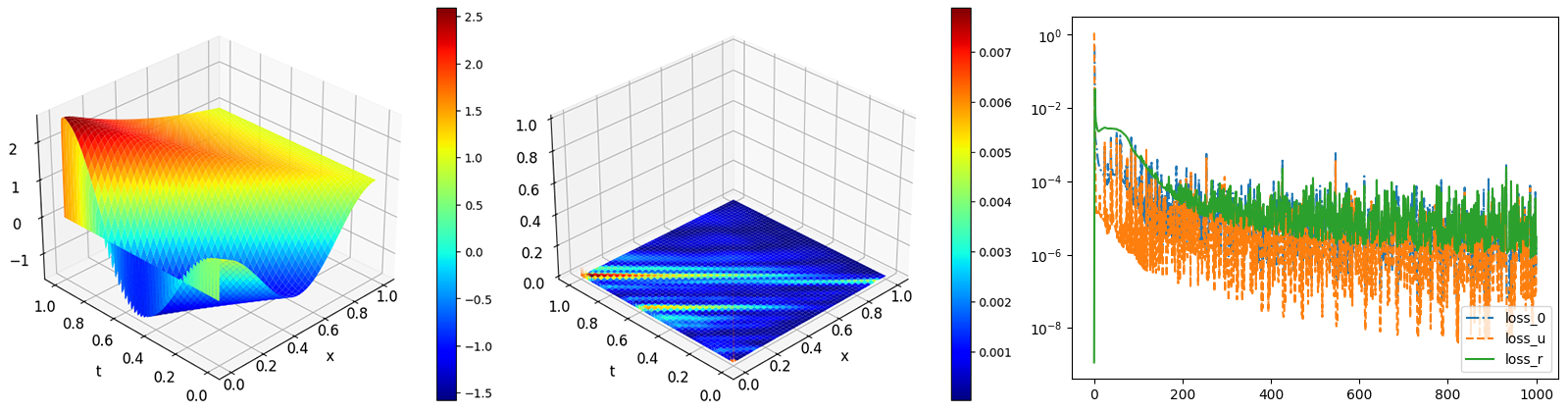} 
        \captionof{subfigure}{GKPINN(Chebyshev-KAN)}
    \end{minipage}

    
    \begin{minipage}{1.0\textwidth}
        \centering
        \includegraphics[width=1.0\textwidth]{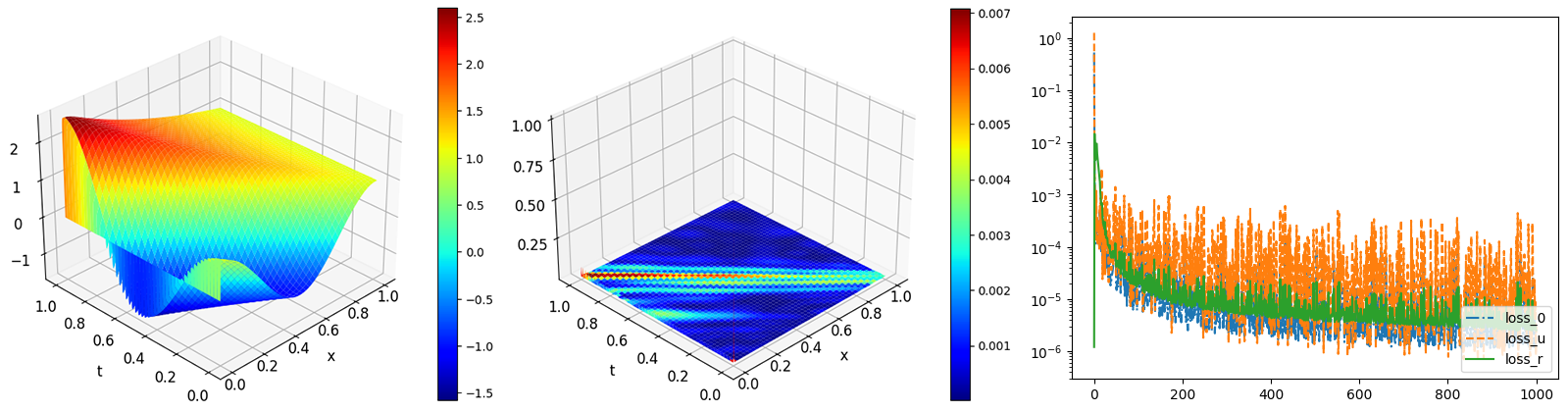}   
        \captionof{subfigure}{ASPINN(MLP)}
    \end{minipage}

    
    \begin{minipage}{1.0\textwidth}
        \centering
        \includegraphics[width=1.0\textwidth]{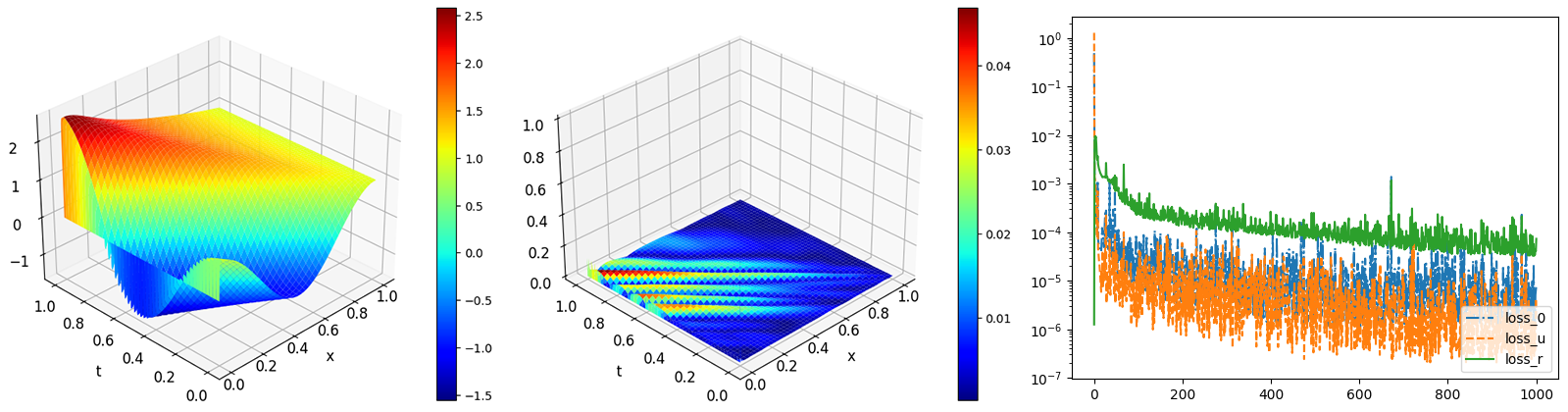}   
        \captionof{subfigure}{ASPINN(Chebyshev-KAN)}
    \end{minipage}
    
    \caption{Solution profile (Left), numerical error (middle), and loss error (right)}
    \label{fig8}
\end{figure}

\begin{figure}[htbp]
    \centering
    \begin{minipage}{1.0\textwidth}
        \centering
        \includegraphics[width=1.0\textwidth]{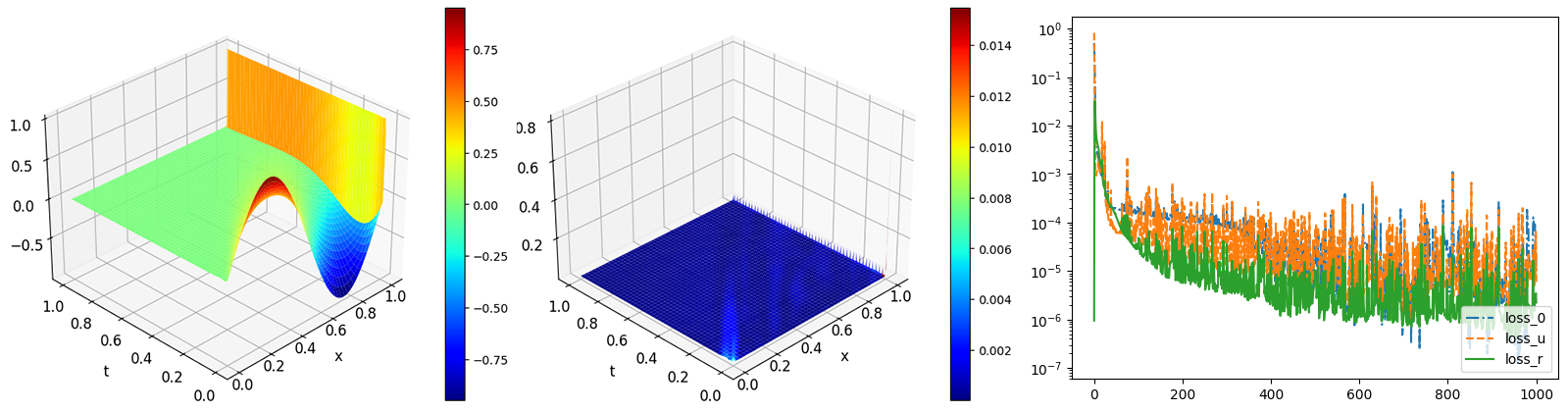}  
        \captionof{subfigure}{GKPINN(MLP)}
    \end{minipage}
    
    
    \begin{minipage}{1.0\textwidth}
        \centering
        \includegraphics[width=1.0\textwidth]{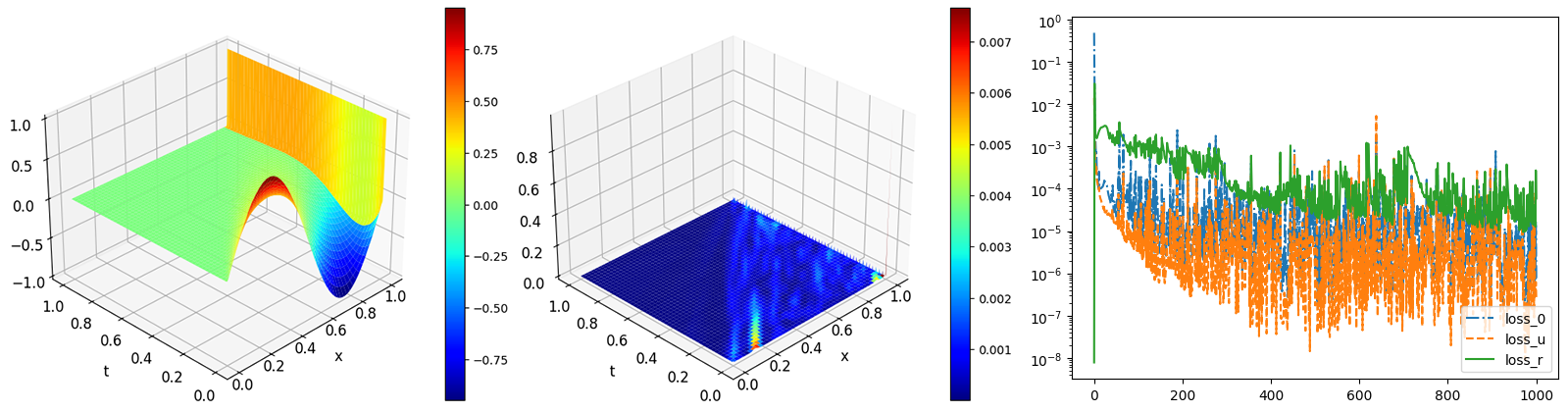} 
        \captionof{subfigure}{GKPINN(Chebyshev-KAN)}
    \end{minipage}

    
    \begin{minipage}{1.0\textwidth}
        \centering
        \includegraphics[width=1.0\textwidth]{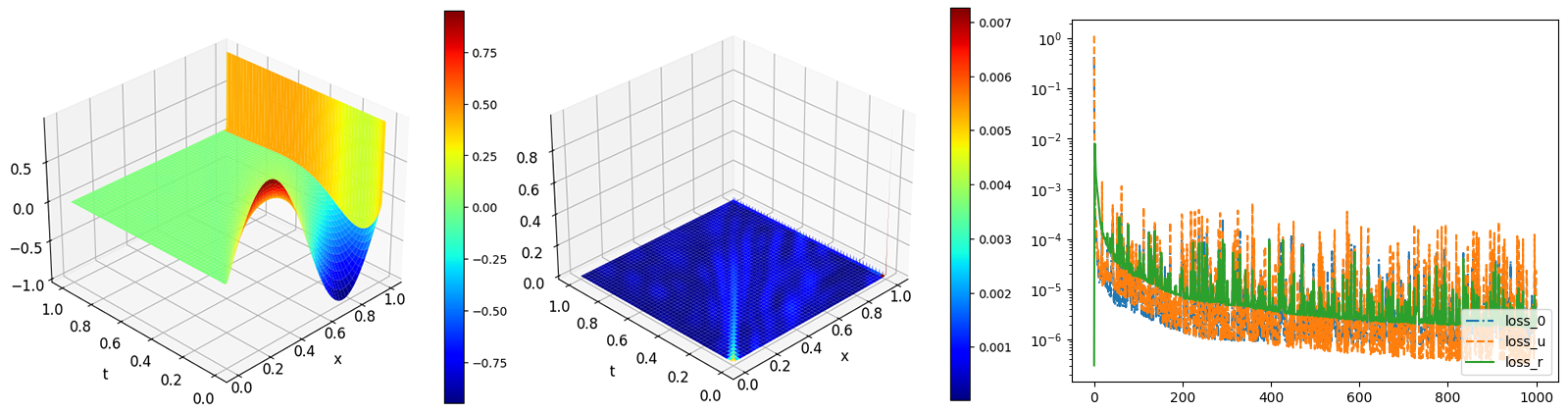}   
        \captionof{subfigure}{ASPINN(MLP)}
    \end{minipage}

    
    \begin{minipage}{1.0\textwidth}
        \centering
        \includegraphics[width=1.0\textwidth]{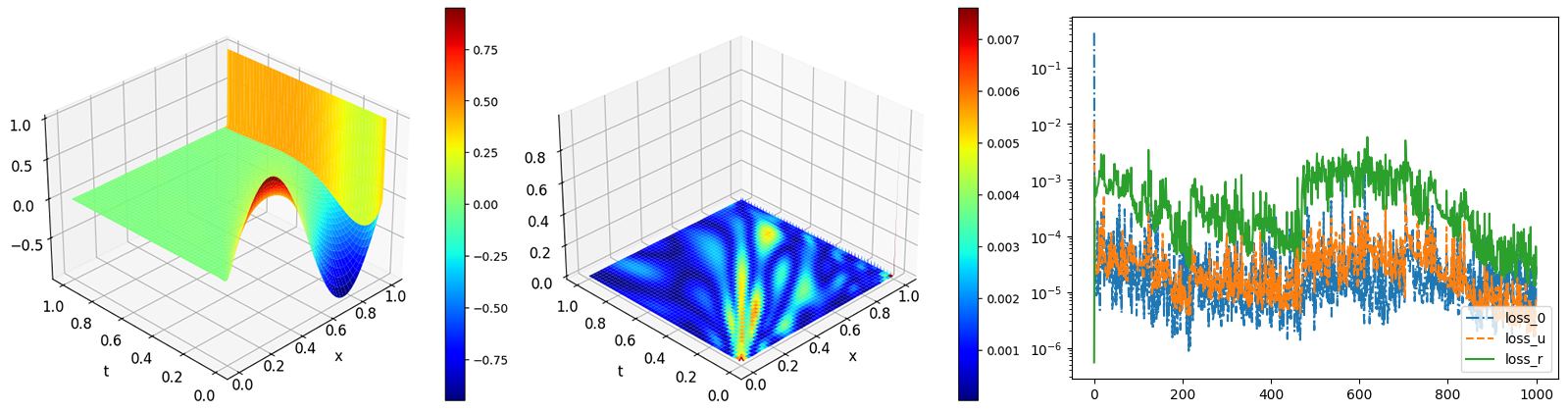}   
        \captionof{subfigure}{ASPINN(Chebyshev-KAN)}
    \end{minipage}
    
    \caption{Solution profile (Left), numerical error (middle), and loss error (right)}
    \label{fig9}
\end{figure}

The asymptotic expansion of $u$ is given by: 
\begin{equation} 
\label{eq42} 
u_{as}(x,t) = u_0(x,t) + (0 - u_0(0,t))e^{\frac{-x}{\varepsilon}}
\end{equation}

As shown in Figure 8, the features of the solutions have a similar effect to being cut at the boundary layer. By observing the loss error plots, it can be found that Chebyshev-KAN pays insufficient attention to $loss_r$, which may be the reason for the poor training effect in Figure 8(d). The exact numerical results can be seen in Table 5:
\begin{table*}[htbp]
\centering
\includegraphics[width=0.5\columnwidth]{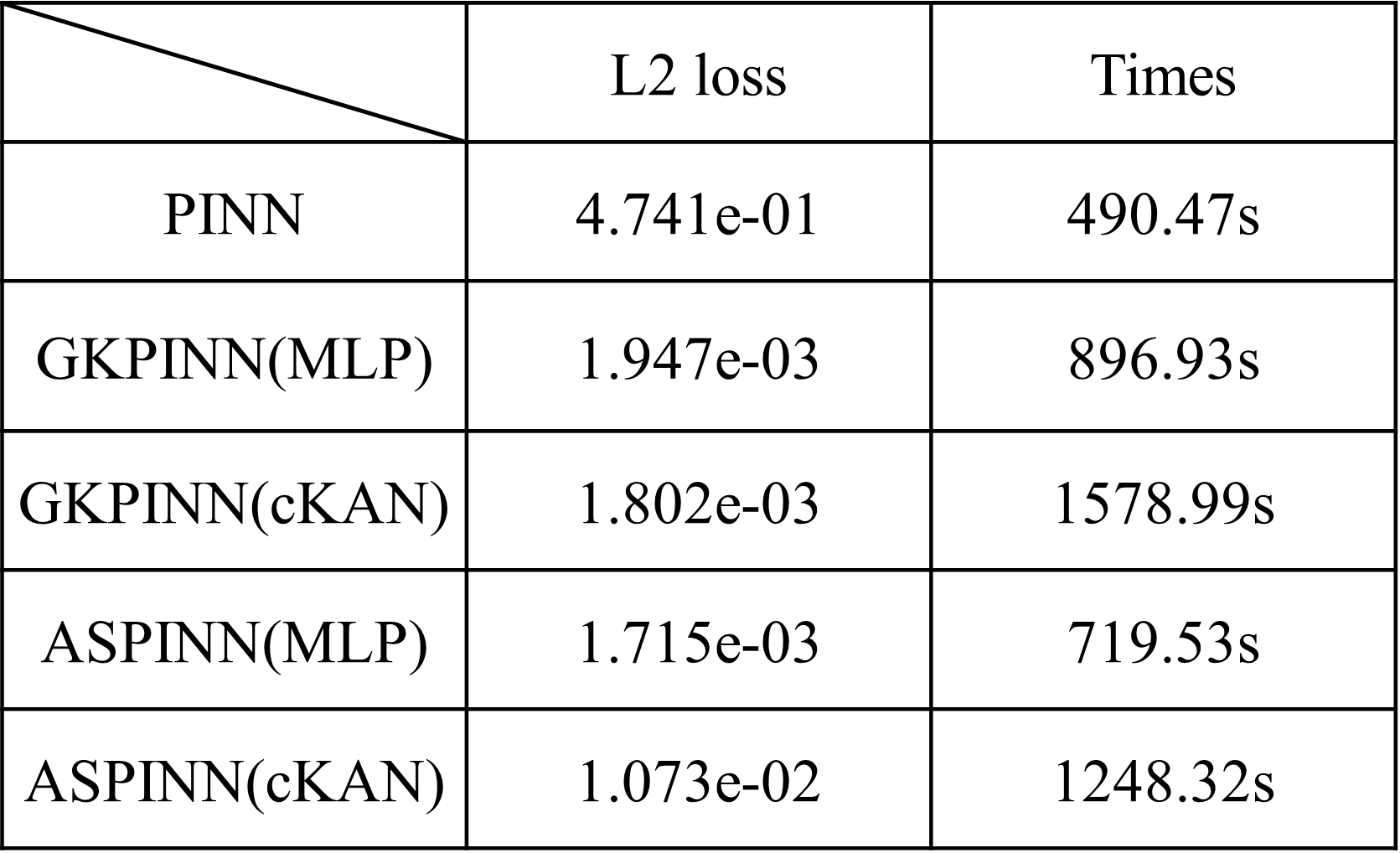} 
\caption{Relative $L_2$ and computational time comparison between different models and training strategies in Eq. (41). Time is measured on Nvidia GeForce RTX-3090 GPU}
\label{table5}
\end{table*}

It can be seen that ASPINN is faster than GKPINN and has a $12\%$ improvement in performance. But Chebyshev-KAN is not as effective as MLP in such problems, whether training efficiency or precision.
\subsubsection{Example 6}
\label{subsec5.2.4}
\begin{equation} 
\label{eq43} 
\begin{cases}
u_t - \varepsilon u_{xx} + u_x +5u = 0, \, (x,t) \in (0,1) \times (0,1] \\ 
u(x,0) = sin(2\pi x) \\
u(0,t) = 0,u(1,t) = 1 \\
\end{cases}
\end{equation}

In this equation, $b = b(x,t) = 1 > 0$, and the boundary layer is positioned at the specific coordinate $x = 1$. We propose the exponential layer as $e^{-\frac{1-x}{\varepsilon}}$, which leads to the asymptotic expansion of the solution: 
\begin{equation} 
\label{eq44} 
u_{as}(x,t) = u_0(x,t) + (1 - u_0(1,t))e^{-\frac{1-x}{\varepsilon}}
\end{equation}

We can obtain the numerical solutions, the numerical error, and the loss
error plots in Figure 9. Figures 9(b) and 9(d) show that Chebyshev-KAN's training of time-varying equations is erratic, and whose performance with ASPINN is unsatisfactory. The results for these methods are detailed in Table 6:
\begin{table*}[htbp]
\centering
\includegraphics[width=0.5\columnwidth]{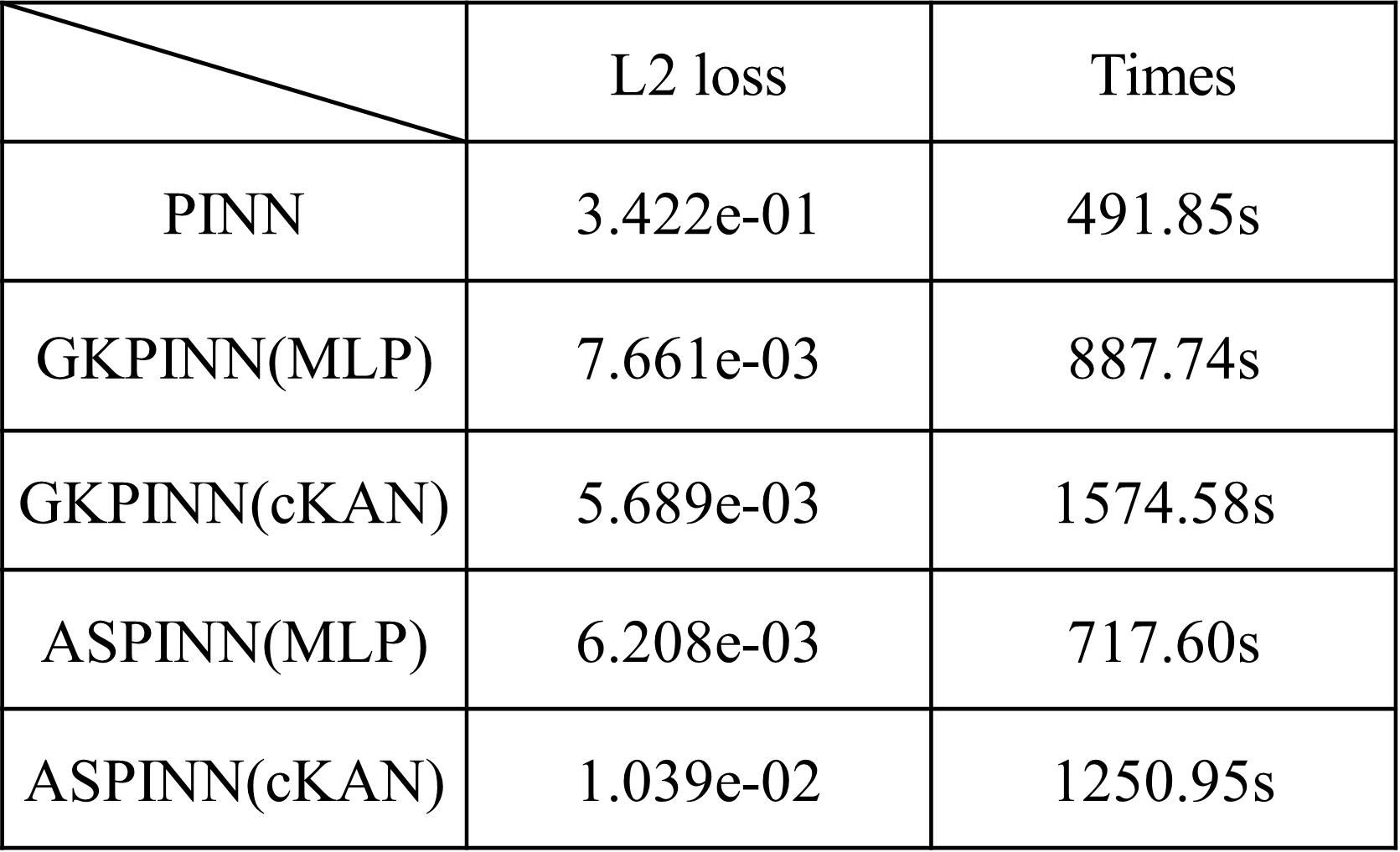} 
\caption{Relative $L_2$ and computational time comparison between different models and training strategies in Eq. (43). Time is measured on Nvidia GeForce RTX-3090 GPU}
\label{table6}
\end{table*}

Overall, when faced with the time-varying equations, the best-performing model is ASPINN+MLP and the effect of Chebyshev-KAN is not ideal.

\section{Conclusion}
\label{sec6}

We propose an Asymptotic Physics-Informed Neural Network for the Singularly Perturbed Differential Equations. This method builds upon the principles of singular perturbation theory and asymptotic expansion methods. The boundary layer part is corrected with the help of the exponential-type layer, and the exact asymptotic expansion of the solution $u$ of the equation is obtained. This strategy is another optimization of the GKPINN architecture. Meanwhile, we introduced Chebyshev-KAN instead of MLP to further improve ASPINN's performance. Among the various SPDEs we have tested, ASPINN shows its advantage in terms of accuracy and training time. Compared to MLP, Chebyshev-KAN achieves superior accuracy, but it takes more GPU hours to train. For SPDEs other than time-varying equations, Chebyshev-KAN is an effective alternative to MLP. For the future research direction, the application scenarios of ASPINN will be extended to the multi-boundary layer problem and the turning point problem, so as to further expand the applicability of the architecture.

\section*{Acknowledgements}
This work was supported by National Key Research and Development Project of China (Grant Nos. 2021YFA1000103, 2021YFA1000102), National Natural Science Foundation of China (Grant Nos. 62272479, 62372469, 62202498), Taishan Scholarship (Grant No. tstq20240506), Shandong Provincial Natural Science Foundation(Grant No. ZR2021QF023).



\end{document}